\documentclass[lettersize,journal]{IEEEtran}
\usepackage{amsmath,amsfonts}
\usepackage{array}
\usepackage[caption=false,font=normalsize,labelfont=sf,textfont=sf]{subfig}
\usepackage{textcomp}
\usepackage{stfloats}
\usepackage{url}
\usepackage{verbatim}
\usepackage{graphicx}
\usepackage{cite}

\usepackage[linesnumbered,ruled]{algorithm2e}
\usepackage{algpseudocode}  
\usepackage{color} 
\usepackage[normalem]{ulem}

\DeclareMathOperator*{\minA}{\,min}

\begin{document}

\makeatletter
\def\ps@IEEEtitlepagestyle{
  \def\@oddfoot{\mycopyrightnotice}
  \def\@evenfoot{}
}
\def\mycopyrightnotice{
  {\footnotesize
  \begin{minipage}{\textwidth}
  \centering
  1051-8215~\copyright~2023 IEEE. Personal use is permitted, but republication/redistribution requires IEEE permission. \\
  See https://www.ieee.org/publications/rights/index.html for more information.
  \end{minipage}
  }
}
 
\title{EC-SfM: Efficient Covisibility-based Structure-from-Motion for Both Sequential and Unordered Images} 

\author{Zhichao~Ye, Chong Bao, Xin Zhou, Haomin Liu, Hujun Bao, \textit{Member, IEEE},\\and Guofeng Zhang,\textit{ Member, IEEE}
\thanks{Manuscript received 9 February 2023; revised 18 May 2023; accepted 27 May 2023. This work was supported in part by NSF of China under Grant 61822310 and Grant 61672457. This article was recommended by Associate Editor X. Cao. (Corresponding author: Guofeng Zhang.)} 
\thanks{Zhichao Ye is with the State Key Laboratory of Computer Aided Design
and Computer Graphics, Zhejiang University, Hangzhou 310058, China,
and also with SenseTime Research, Hangzhou 311215, China (e-mail:
yezhichao\_cad@zju.edu.cn).}
\thanks{Chong Bao, Xin Zhou, Hujun Bao, and Guofeng Zhang are with the State
Key Laboratory of Computer Aided Design and Computer Graphics, Zhejiang
University, Hangzhou 310058, China, and also with ZJU-SenseTime Joint
Lab of 3D Vision (e-mail: chongbao@zju.edu.cn; zhouxin\_cad@zju.edu.cn;
baohujun@zju.edu.cn; zhangguofeng@zju.edu.cn).}
\thanks{Haomin Liu is with SenseTime Research, Hangzhou 311215, China (e-mail:
liuhaomin@sensetime.com).}
\thanks{Digital Object Identifier 10.1109/TCSVT.2023.3285479}
}

 
\maketitle

\begin{abstract}
Structure-from-Motion is a technology used to obtain scene structure through image collection, which is a fundamental problem in computer vision.
For unordered Internet images, SfM is very slow due to the lack of prior knowledge about image overlap.
For sequential images, knowing the large overlap between adjacent frames, SfM can adopt a variety of acceleration strategies, which are only applicable to sequential data.
To further improve the reconstruction efficiency and break the gap of strategies between these two kinds of data, this paper presents an efficient covisibility-based incremental SfM.
Different from previous methods, we exploit covisibility and registration dependency to describe the image connection which is suitable to any kind of data.
Based on this general image connection, we propose a unified framework to efficiently reconstruct sequential images, unordered images, and the mixture of these two.
Experiments on the unordered images and mixed data verify the effectiveness of the proposed method, which is three times faster than the state-of-the-art on feature matching, and an order of magnitude faster on reconstruction without sacrificing the accuracy.
The source code is publicly available at https://github.com/openxrlab/xrsfm.
\end{abstract}

\begin{IEEEkeywords}
structure from motion, covisibility, epipolar geometry, keyframe.
\end{IEEEkeywords}
  
\section{Introduction} 
\IEEEPARstart{O}{ver} the past decades, Structure-from-Motion (SfM) has been an important topic in the field of 3D vision.
Thanks to the robustness of SfM, accurate camera poses and a point cloud model of the scene can be estimated by merely photo collections.
This kind of demand is common in autonomous driving, augmented reality, and other diverse 3D vision applications.
Traditional SfM systems~\cite{rome,lSfM,colmap,moulon2016openmvg,openmvs2020} can reconstruct the scene from the unordered Internet images, but is slow and requires a lot of computing resources.
The common acceleration methods~\cite{enft, resch2015scalable} leverage the image order to save calculation, so that a large amount of data can be processed efficiently. But these methods are only suitable for sequential images as input. 

At present, an increasing number of applications require reconstruction algorithms that support various types of data as input.
For example, city-level crowdsourced reconstruction always faces various data including vehicle-mounted video, Unmanned Aerial Vehicle (UAV) imagery, and street view pictures.
Another example is the use of Internet data to reconstruct famous landmarks.
In the past, landmark reconstruction often used Internet photo collections. 
Now, with the development of video websites, rich Internet videos can also be used.
The mixture of unordered and sequential images brings new challenges to SfM.
Dealing with large-scale mixed data, existing unordered strategies bear the huge computational burden, and the sequential strategies are not suitable for unordered data, requiring a new SfM method that can reconstruct from mixed data accurately, efficiently, and completely.

\begin{figure}
\centering
\includegraphics[width=1.0\linewidth]{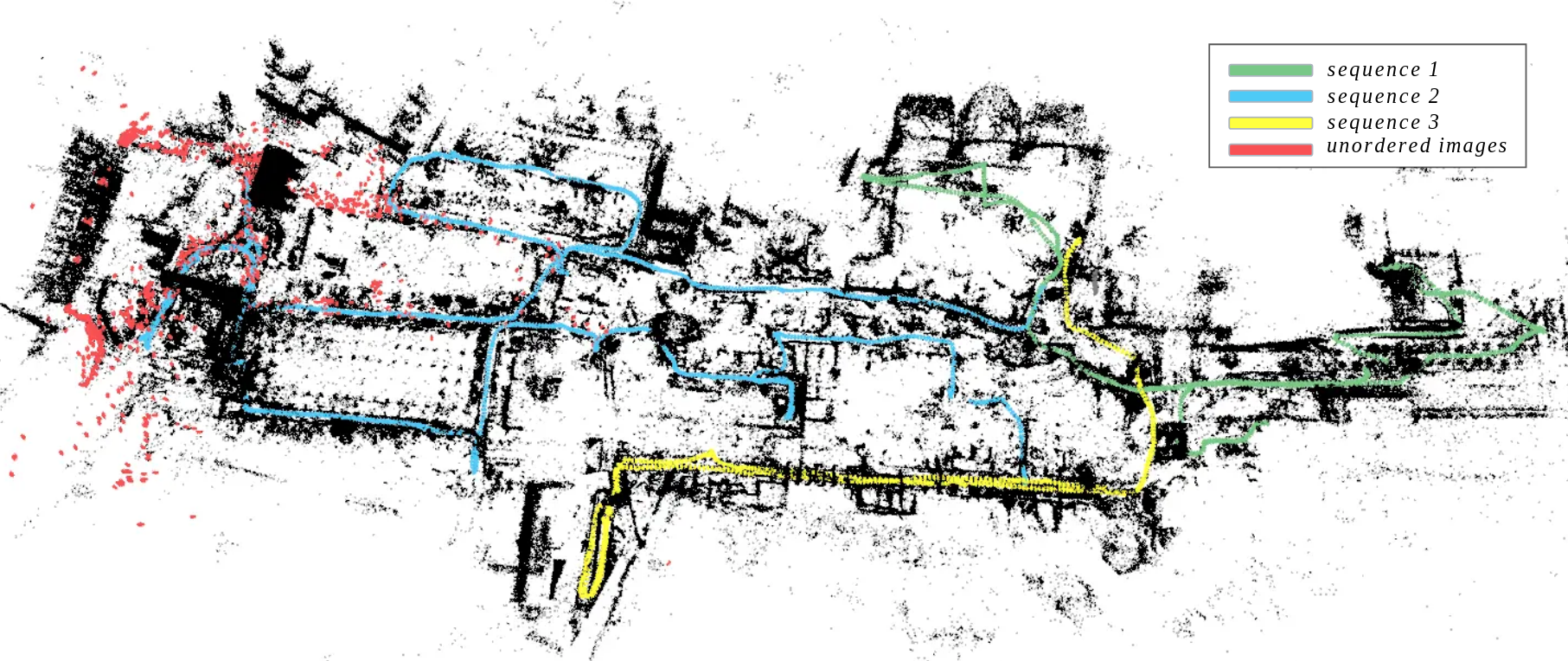}
\caption{The reconstruction result of Roman Forum with the proposed SfM system. Different colors distinguish different input data, including multiple image sequences and Internet images.}
\label{fig:Rome mix result}
\end{figure}

We find that the essence of the sequential strategies is to reduce the redundant matching and optimization with knowing the large overlap of adjacent frames in sequential data.
Inspired by this, we extract covisibility relations and registration dependencies from images to better describe the internal relationship in various data. 
By utilizing these internal relationship, much time wasted in redundant calculations is saved, and the internal relationship is suitable for both sequential and unordered images.
In this paper, we propose a covisibility-based incremental SfM system,  which uses a unified framework to efficiently process sequential images, unordered images, and mixed data.
The proposed SfM system is much faster than the traditional SfM systems for unordered images and mixed data.
As shown in Fig.~\ref{fig:Rome mix result}, the Roman Forum was completely reconstructed in half an hour with three video sequences and an Internet photo collection using the proposed SfM system.

The preliminary conference version~\cite{ecim} of this work only focused on accelerating the matching process of SfM by leveraging the covisibility information of unordered images.
In this paper, we extend it with an image clustering strategy to ensure that the algorithm can also run efficiently on sequential images. 
Besides, we design a complete SfM system by proposing a novelty reconstruction framework that can efficiently process the mixture of sequential and unordered images.  
Reviewing the reconstruction stage of traditional incremental SfM, we note that the robustness of most systems~\cite{rome, lSfM, colmap} depends on frequent bundling adjustments to suppress cumulative errors, but it brings a heavy computational burden.
Moreover, due to the accumulated error, some correct 2D-3D matching may be considered as outliers,
so that the loop cannot be closed.
In SLAM systems, keyframe selection~\cite{dong2009keyframe,holmes2009relative} and loop closure~\cite{strasdat2010scale,mur2017visual} are common modules, which can solve the above problems, but they are only suitable for sequential images.
In order to adapt to mixed data, we propose a keyframe selection method based on registration dependency and a new geometric verification algorithm for covisible image pairs.
This greatly improves the reconstruction speed and is suitable for sequential images and unordered images.

To sum up, our major contributions are as follows:
 \begin{itemize}
	\item We propose a powerful SfM system that can handle various data types in a unified framework, including sequential images, Internet photo collection, and mixed data.
	
	\item We propose a covisibility-based matching strategy that discovers covisible image pairs and iteratively extends the feature matches from the potential registration images.
	
	\item We proposed a hierarchy-based keyframe selection method to speed up reconstruction and an error detection method to close loops. 
	The proposed method is not limited to sequential images, and can process well on unordered images and mixed data.

    \item Experiments on the Internet photo collection and mixed data verify the effectiveness of the proposed method, which is three times faster than the state-of-the-art on feature matching, and an order of magnitude faster on reconstruction, without sacrificing the accuracy.
\end{itemize}

\section{Related Work}
The SfM technique has achieved great success in the past decades~\cite{rome, lSfM, enft, colmap, global,zhu2018very,moustakas2005stereoscopic,gao2019complete}.
The general pipeline of SfM contains two major stages: the matching stage and the reconstruction stage.
The matching stage mainly carries out feature extraction and matching. 
The reconstruction phase is responsible for estimating the camera poses and map points from the feature matches.
We review the two stages in this section.

\subsection{Matching Stage}
To find feature correspondences among the whole image set, the most straightforward strategy is performing feature matching between each image pair, which is infeasible for a large image set.
Image retrieval techniques~\cite{image_retrival} can be used to find candidate image pairs for further feature matching.
Vocabulary tree~\cite{vocabularytree} is a representative image retrieval method, which is widely used in various SLAM and SfM systems~\cite{orbslam,bow_based_sysytem}.
This method clusters features to build a visual dictionary, and uses the distribution of words to compute similarities between images.
Other kinds of image retrieval methods use global descriptors instead of local descriptor sets to represent images, such as GIST~\cite{GIST}.
With the great success of deep learning in computer vision, image retrieval methods based on Convolutional Neural Networks~(CNNs)~\cite{netvlad} have emerged. 
These learning-based methods have a stronger image representing ability, that are more robust to changes in illumination and view point.

After image retrieval, the common strategy is matching the $N_R$ most similar image for each image.
However, it is difficult to decide the fixed number of $N_R$ before actually performing feature matching. 
Depending on the capture density, camera field of view, scene distance, and many other factors, some images would have many overlapping pairs and others would have few. 
Using a fixed number of $N_R$ easily leads to a lack of feature matches resulting in incomplete reconstruction or waste of computation for matching image pairs without any common features.
A simple improvement method~\cite{bundler} uses query expansion, which matches the query results of neighbor frames. 
This method can obtain some missing matches, but costs a lot for images with rich matching relations.
Another method is MatchMiner~\cite{lou2012matchminer}, 
which is based on the vocabulary tree, using weights to distinguish valuable vocabularies from noisy vocabularies to achieve better performance.
VocMatch~\cite{vocmatch} improves the vocabulary tree algorithm by considering features indexed to the same visual word as potential matches to skip the descriptor matching. 
ENFT~\cite{enft} proposes to construct a matching matrix and select frame pairs with the maximum overlapping confidence for feature matching, and use the matching result to update the matching matrix iteratively.
Finally, a vote-and-verify strategy~\cite{vav} of vocabulary tree was proposed for fast spatial verification.

The covisibility graph is the data structure to represent the image matching relationship. 
In the covisibility graph, images are represented as nodes, and the edge between two nodes indicates there are common features between the image pair. 
ORB-SLAM~\cite{orbslam} builds a covisibility graph to efficiently find candidate keyframes to be matched with the current frame on sequential images.
Mei \textit{et al.}~\cite{covisibility} use a similar idea to handle place recognition and loop closure.
In our previous work\cite{ecim}, we used image retrieval with a small $N_R$ to get initial feature matches and construct the covisibility graph for unordered image set, and leveraged the transitivity of covisibility to predict overlapping images and extend feature matches in the iterative manner. 
In this paper, we extend this method to the mixture of sequential and unordered images.

With the development of deep learning, many learning-based retrieval~\cite{netvlad} and feature methods~\cite{detone2018superpoint,sarlin2020superglue,wang2020displacement} have been proposed.
These learning-based methods can be easily integrated with the proposed matching method which use only the covisibility and do not limit the use of features and retrieval methods.

\subsection{Reconstruction Stage} 
After years of development, the reconstruction stage of SfM has made great progress. 
Currently, reconstruction methods can be categorized into incremental SfM and global SfM.

Incremental SfM builds the initial map using two-view reconstruction~\cite{brown1958solution}.
And then, the poses of cameras that observed enough map points in the initial map can be estimated. 
After a camera pose is registered, the map is extended through triangulation.
The complete scene structure is constructed by iteratively estimating camera poses and extending the map.
The first incremental system for Internet photo collection is proposed by Snavely \textit{et al.}~\cite{phototourism} with exhaustive image pair matching and frequent call for bundle adjustment (BA).
As an attempt to scale to a large photo collection, \cite{snavely2008skeletal} exploited the skeleton graph of the scene, and \cite{li2008modeling, frahm2010building} highlighted the iconic image representing the main structure of the scene graph.
Wu proposed a linear-time SfM~\cite{lSfM}, which introduces the preconditioned conjugate gradient and adjusts the optimization frequency to make the optimization time linear.
The current state-of-the-art SfM system COLMAP\cite{colmap} developed several well-designed strategies to further improve the reconstruction quality, such as the multi-model geometric verification to enhance robustness, visibility pyramid to select next best view, and an iterative BA, re-triangulation and outlier filtering to enhance the robustness of the system.
In addition, OpenMVG~\cite{moulon2016openmvg} and OpenMVS~\cite{openmvs2020} are two well-known open-source frameworks. 
In recent years, many deep learning-based reconstruction methods~\cite{vijayanarasimhan2017sfm,zhou2017unsupervised,yin2018geonet,zou2018df,wei2020deepsfm,wang2021deep,cai2021extreme,liu2022depth,sarlin2021back} have emerged. 
Among them, SfM-Net~\cite{vijayanarasimhan2017sfm} and GeoNet~\cite{yin2018geonet} use neural networks to simultaneously predict image depth, camera motion, and optical flow, which introduce more observational information and optimize it through geometric constraints.
Wang \textit{et al.}~\cite{wang2021deep} adopt a similar approach, but focus on alleviating the ill-posedness problem in two-view reconstruction.
Other researchers make efforts to improve the robustness of geometric estimation under extreme or special circumstances~\cite{cai2021extreme,liu2022depth}.
In addition, Sarlin \textit{et al.}~\cite{sarlin2021back} focus on utilizing multi-view information and geometric constraints to improve the detection accuracy of feature points through inverse optimization.

Different from incremental SfM, global SfM recovers all camera poses in the batch manner. 
All camera poses are initialized by motion averaging and refined by global optimization.
As the core of global SfM, the majority of motion averaging methods estimate rotation and translation separately.
The early works of rotation averaging~\cite{govindu2004lie} solve the problem by linear least squares.
In order to reduce the influence of outliers, researchers adopt Iteratively Reweighted Least Squares~\cite{chatterjee2017robust} and regularization terms~\cite{crandall2012sfm} to enhance the robustness.
Gao \textit{et al.}~\cite{gao2021ira++,gao2022irav3} estimate absolute rotations in an incremental manner to obtain accurate camera orientations.
Besides, \cite{chen2021hybrid} adopted a hybrid method that combines a global optimizer and local optimizer to gain outlier resistance.
The translation estimation methods can be roughly divided into essential matrix based methods \cite{ brand2004spectral,arie2012global, ozyesil2015robust} and Trifocal tensor based methods\cite{sim2006recovering, courchay2010exploiting, jiang2013global}.
In order to further improve the efficiency of global SfM, \cite{zhu2018very} proposes a divide-and-conquer framework to realize city-level reconstruction.
\cite{cai2021pose} propose a pose-only reconstruction method that gives a linear global translation solution and represents 3D points by camera parameters in the optimization so that the efficiency is greatly improved.

Compared with incremental SfM, global SfM avoids frequent calls for the time-consuming BA and also alleviates the risk of error accumulation, but is still very sensitive to outliers. 
There are hybrid methods\cite{bhowmick2014divide, gao2022incremental} that adopt the compromise scheme, utilizing the global rotation averaging and the incremental translation estimation to balance the efficiency and robustness. 
 
However, due to the superior robustness, incremental SfM is still the mainstream of the reconstruction system, and our method also falls into this category.
Based on the traditional incremental reconstruction systems, we proposed the hierarchy-based keyframe selection and error correction module, which greatly improves the reconstruction speed and is suitable for any type of data.

\section{Overview}

\begin{figure}
	\centering
	\includegraphics[width=1.0\linewidth]{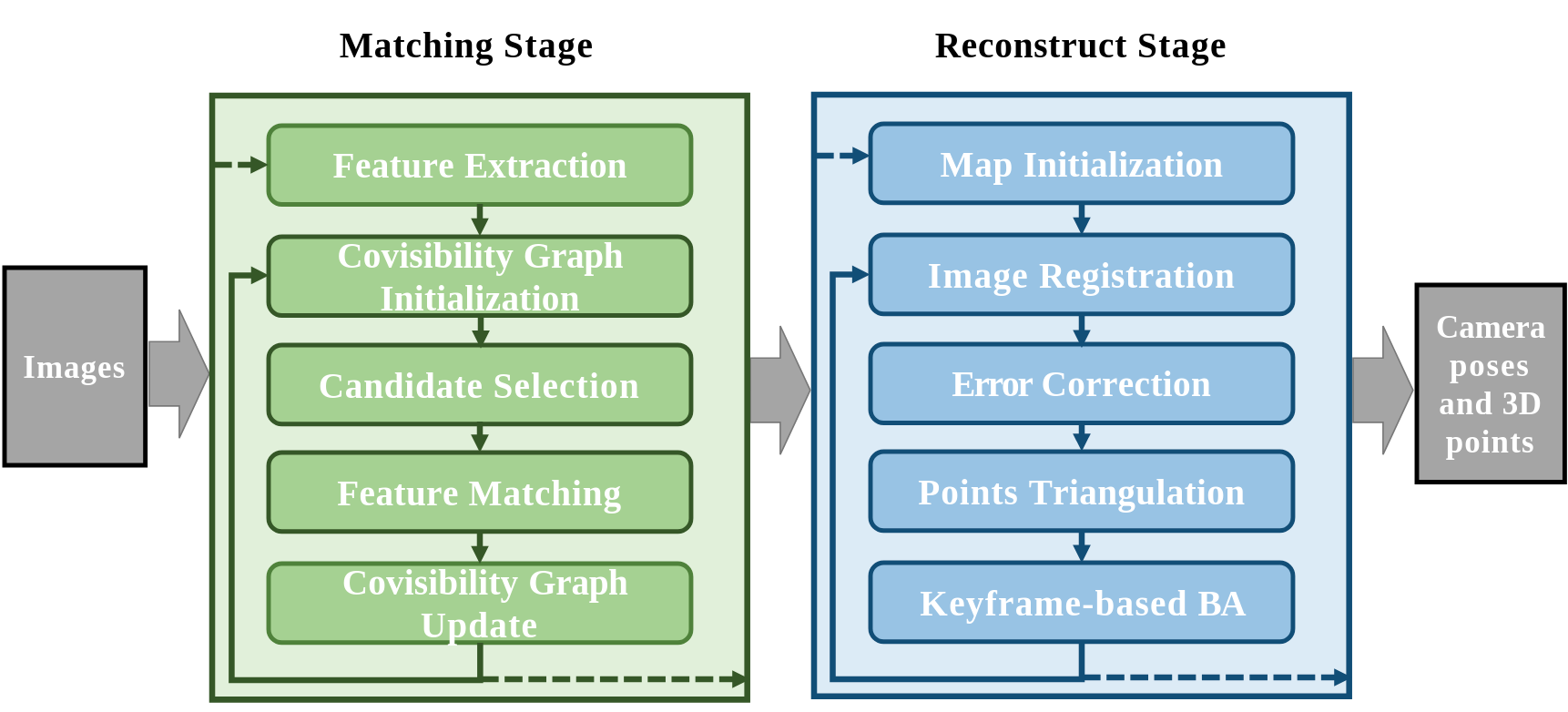}
	\caption{The framework of the proposed method.}
	\label{fig:Sys-pipeline}
\end{figure}

The framework of the proposed method is shown in Fig.~\ref{fig:Sys-pipeline}, which is comprised of the matching stage and the reconstruction stage.
In the matching stage, we first extract features for each image and construct an initial covisibility graph.
Then, based on the covisibility graph, we select the covisible image pairs as the candidates for feature matching.
The whole matching process is iterative, the results of each round of feature matching are used to update the covisibility graph to search for new covisible image pairs and more feature matches.
In the reconstruction stage, similar to the traditional incremental SfM, we have an initialization module and iteratively perform image registration and triangulation.
To handle the mixed data of sequential and unordered images, we propose an error correction method to solve the loop closure problem and a keyframe-based BA strategy to improve efficiency.

\section{Matching Stage}
\begin{figure*}
\centering
\includegraphics[width=0.9\linewidth]{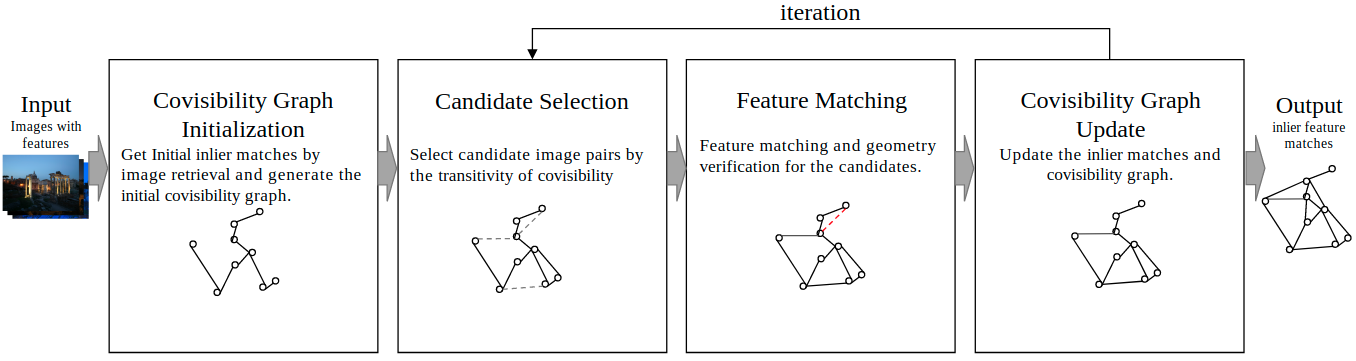}
\caption{The pipeline of our iterative matching strategy.}
\label{fig:matching-pipeline}
\end{figure*}

This section introduces the proposed matching method.
Different from previous methods, we propose the region-based covisibility graph and leverage the transitivity of covisibility to find potential overlapping image pairs.
In addition, we adopt the iterative matching strategy that extends matches from high-quality images (potential registration frames) to avoid matching between low-quality images.

\subsection{Region-based Covisibility Graph}

The matching stage searches image correspondences among the input image set $U = \{ I_i \mid i=1...N_I \}$.
A set of features are extracted from each image $i$, denoted as $\mathcal{F}_i = \{ f_i^k \mid k=1...N_{\mathcal{F}_i} \}$, where $f_i^k$ denotes the $k$-th feature in the $i$-th image, and $N_{\mathcal{F}_i}$ is the number of features.
Next, the features correspondence is established by feature matching.
If we have a part of feature matches in hand, we can predict the potential covisibility of other images.
For example, given matched image pairs $(I_a,I_b)$ and $(I_b,I_c)$, if $I_a$ and $I_c$ share some feature tracks,  $(I_a,I_c)$ is also a covisible image pair. 
Unfortunately, there are inevitably some mismatches in the feature matching result even after the geometric verification, which makes some covisibility predictions unreliable.
In addition, the projections of 3D scene points sometimes are not detected so that some covisible image pairs would be missed.
Predicting covisibility directly from feature matches is greatly affected by missing matches and mismatches.

\begin{figure}
\centering
\includegraphics[width=1.0\linewidth]{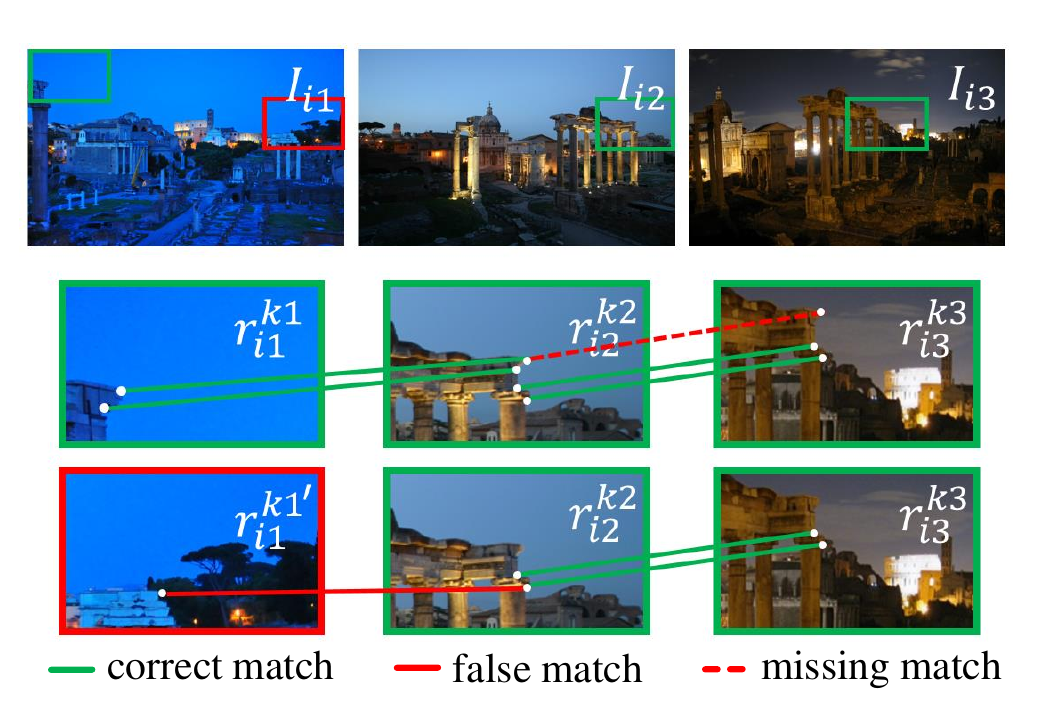}
\caption{The top row contains three covisible images; the second row shows the correct transitive covisibility ~($r_{i1}^{k1}$, $r_{i3}^{k3}$); In the third row, ($r_{i1}^{k1'}$, $r_{i3}^{k3}$) shares a feature track, but it is not a covisible region pair.
}  
\label{fig:covisibility} 
\end{figure}

We propose a region-based covisibility prediction method to address the problems of missing matches and mismatches.
Due to the existence of mismatches, covisibility prediction using feature points is not reliable. 
Instead, we use the local region where the feature is located to construct a flexible covisibility of regions.
Through region-based covisibility, covisible image pairs that do not share the feature track due to missing matches may also be found. 
As shown in Fig.~\ref{fig:covisibility}, there is no feature track shared by $r_{i_1}$ and $r_{i_3}$ due to missing matches, 
but the potential covisible relation between $r_{i_1}$ and $r_{i_3}$ can be found by the transitivity of region-based covisibility,
because both $(r_{i_1}^{k_1},r_{i_2}^{k_2})$ and $(r_{i_2}^{k_2},r_{i_3}^{k_3})$ are covisible region pairs.
Moreover, in order to prevent the false-positive detection of covisible region pairs supported by very few feature tracks, we use the number of shared feature tracks to measure the confidence of covisibility. 
The covisibility of the region pairs, e.g., $(r_{i_1}^{k_1},r_{i_2}^{k_2})$ and $(r_{i_2}^{k_2},r_{i_3}^{k_3})$, are relatively reliable because there are supported by sufficient feature tracks.
By contrast, the region pair $(r_{i_1}^{k_1'},r_{i_2}^{k_2})$ shares only one feature track, so this pair is not considered as a covisible pair.

For characterize region-based covisibility efficiently, we uniformly divide each image $I_i$ into $N_p \times N_p$ patches ($P_i=\{p_i^{k}|k = 1...N_p^2\}$) as a region approximation. 
A patch pair $(p_i^{k1}, p_j^{k2})$ is considered covisible if they share at least $T$ common feature tracks.
And we form a region-based covisibility graph with patches as nodes and covisible patch pairs as edges.
For two images $I_i$ and $I_j$, if there is a chain $(p_{1},p_{2},...p_{n})$ in the region-based covisibility graph, where $p_{1}$ belongs to $I_i$ and $p_{n}$ belongs to $I_j$, then these two images are potentially covisible.
However, for a long chain $(p_{1},p_{2},...p_{n})$, there is a large overlap between every $p_{i},p_{i+1}$, but with the increase of chain length, the overlap between $p_{1}$ and $p_{n}$ becomes small or even completely disappeared, and the predicted covisibility is unreliable.
Therefore, to avoid this situation, we set a length threshold $\sigma$ for the maximum allowed chain length.

Specifically, we define the distance between an image pair from the covisibility of patches by 
\begin{equation}
dist(I_i,I_j) = \minA_{} \{dist(p_i^{k_1},p_j^{k_2})|k_1\in[1,N_P^2],k_2\in[1,N_P^2]\},
\label{Eq:patch_graph}
\end{equation}
where $dist(p_i^{k1},p_j^{k2})$ is the distance between $p_i^{k1}$ and $p_j^{k2}$ in the region-based covisibility graph.
If there is no path connecting the two patches, $dist(p_i^{k1},p_j^{k2})$ is defined as infinite.
$dist(I_i, I_j)$ is the minimum distance between patches in $I_i$ and in $I_j$. 
We detect the covisibility between $I_i$ and $I_j$ by the following equation:
\begin{equation}
covisible(I_i,I_j)=\left\{
\begin{aligned}
True & &\ dist(I_i,I_j) < \sigma\\
False & &\ dist(I_i,I_j) \geq \sigma\\
\end{aligned},
\right.
\label{Eq:covisibility}
\end{equation}
here, $I_i$ and $I_j$ are regarded as covisible if $dist(I_i, I_j)$ does not exceed the threshold $\sigma$.
In our implementation, $\sigma$ is set to 3 for reliable covisibility prediction.

\subsection{Iterative Matching Strategy}

\begin{figure}
\centering
\includegraphics[width=0.9\linewidth]{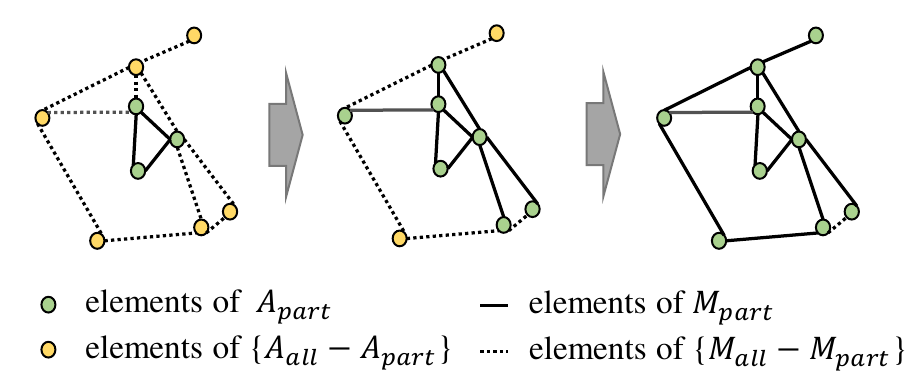}
\caption{Iteratively extending the registered images.}
\label{fig:iterative-extend} 
\end{figure}
 
We propose an iterative algorithm that make full use of existing feature matches to extends feature matches, as illustrated in Fig.~\ref{fig:matching-pipeline}. 
Firstly, we use image retrieval for each image and perform feature matching on the $N_{init}$ most similar images, $N_{init}$ is set to 5 in our implementation.
The set of the initial feature matches is denoted as $\mathcal{M}_{init}$, and a region-based covisibility graph is built with $\mathcal{M}_{init}$.
At each iteration, we select and match all the covisible image pairs based on Equation (\ref{Eq:covisibility}) to obtain more feature matches. 
The newly matched features are used to update the region-based covisibility graph.
In this way, most of the feature matches can be found after several iterations.
 
In addition, we observe there are many poor quality or irrelevant images in Internet photo collections. 
These images cannot be registered successfully even if all the feature matches have been established.
Let $\mathcal{M}_{all}$ denote all the feature matches, $A_{reg}$ denote the set of images that can be registered with the support of $\mathcal{M}_{all}$, and $A_{rest}$ denotes the rest. 
To further reduce invalid matches, we want to avoid selecting candidate image pairs composed of elements in $A_{rest}$.
However, $A_{reg}$ is impossible to be obtained until $\mathcal{M}_{all}$ is obtained and reconstruction is implemented.
In each iteration of the proposed method, we have partial feature matches $\mathcal{M}_{inlier}$ (a subset of $\mathcal{M}_{all}$), so partial images $A_{part}$ can be registered. 

For obtaining $A_{part}$, a naive method is running the reconstruction with existing feature matches $\mathcal{M}_{inlier}$, but it causes a strong coupling between the reconstruction process and matching process. 
Therefore, we propose a fast algorithm that simulates the registration process to obtain an approximation of $A_{part}$. 
Before diving into details, we explain some definitions.

\begin{equation}
Tri(I_i,I_j)=\left\{
\begin{aligned}
	f_i^a,f_j^b|(f_i^a,f_j^b) \in \mathcal{M}_{inlier}\}
\end{aligned}
\right.
\label{Eq:tri}
\end{equation}

\begin{equation}
Match(I_i,\mathcal{F}) =\left\{
\begin{aligned}
	(f_i^a,f_j^b)|(f_i^a,f_j^b) \in \mathcal{M}_{inlier}, f_j^b \in \mathcal{F} \}
\end{aligned}
\right.
\label{Eq:match}
\end{equation} 

The matched features $f_i^a$ and $f_j^b$ between image pair $(I_i,I_j)$ make up the point set $Tri(I_i,I_j)$ for simulating the progress of triangulation in reconstruction. Given a feature set $\mathcal{F}$, $Match(I_i,\mathcal{F})$ is the set of feature matches between features of $I_i$ and $\mathcal{F}$, which is used to detect the number of points in $\mathcal{F}$ observed by $I_i$ for simulating the process of pose estimation.

\begin{algorithm}[t]
\caption{Registration approximation algorithm} \label{alg:approximation}

\LinesNumbered 
\KwIn{inlier matches $ \mathcal{M}_{inlier}$, threshold $t$} 
\KwOut{$A_{appr}$} 

$A_{appr} = \emptyset$, $\mathcal{F} = \emptyset$;\label{init1}

Select a matched image pair $(I_i,I_j)$.\label{init2}

$A_{appr} = A_{appr} \cup \{I_i,I_j\}$\;\label{init3}

$\mathcal{F} = \mathcal{F} \cup Tri(I_i,I_j)$\;\label{init4}

$found \gets True$;\label{iterate_stage_start}

\While{$ found $}
{
	$found \gets False$\;
	\For{$I_i \in U-A_{appr}$}
	{
		\If{$| Match(I_i,\mathcal{F}) |> t$ }
		{
			$A_{appr} \gets A_{appr}\cup\{I_i\}$\;
			$found \gets True$\;
			\For{$I_j \in A_{appr}$}
			{
				$\mathcal{F} \gets \mathcal{F} \cup Tri(I_i,I_j)$\;
			}
		}
	}
} \label{iterate_stage_end}
\end{algorithm} 

\begin{algorithm} [t]
\caption{Iterative matching strategy} \label{alg:iteration}

\LinesNumbered 
\KwIn{initial inlier matches $\mathcal{M}_{init}$, retrieval param $N_{max}$} 
\KwOut{inlier matches $\mathcal{M}_{inlier}$} 

$ \mathcal{M}_{inlier}$ = $ \mathcal{M}_{init}$\;

$C = \emptyset$\;

Compute $A_{appr}$ from $ \mathcal{M}_{inlier}$ by 
Algorithm~\ref{alg:approximation}.

\For{$I_i \in A_{appr}$}
{
	\For{$I_j \in Retrieval(I_i,N_{max})$ \label{retrieval}} 
	{
		\If{$covisible(I_i$,$I_j)$}
		{
			$C \gets C \cup {(I_i,I_j)}$\;
		}
	} 
}

Verify the candidate pairs in $C$ and update $ \mathcal{M}_{inlier}$.

Repeat from line 2 until the maximum number of iterations is reached.
\end{algorithm} 

In order to obtain the potential registration frame set, we select two frames as initial potential registration frames, and then iterate the extended feature set $\mathcal{F}$ and the potential registration frame set $A_{appr}$.
$A_{appr}$ is an approximation of $A_{part}$. 
As shown in Algorithm ~\ref{alg:approximation}, our registration approximation algorithm includes an initial stage (step~\ref{init1} to~\ref{init4}) and an iteration extending stage (step~\ref{iterate_stage_start} to~\ref{iterate_stage_end}), which is similar to the reconstruction stage of SfM.

Combining region-based covisibility graph and the registration approximation algorithm, the final matching algorithm is shown in Algorithm~\ref{alg:iteration}.
Firstly, we build the region-based covisibility graph $G_{cov}$ and obtain the potential registration frames $A_{appr}$ with the initial inlier matches.
Then, we iteratively select the candidate image pairs to match with $G_{cov}$ and $A_{appr}$, and update $A_{appr}$ and $G_{cov}$ with the matching results.
To limit the matching time, the image pairs that need to be tested by covisibility are selected only from retrieval results(step ~\ref{retrieval}).
$Retrieval(I_i,N_{max})$ denotes the retrieval results of $I_i$ with the retrieval number $N_{max}$.
Because $I_j$ is limited in $Retrieval(I_i,N_{max})$, in the worst case, the proposed method will match each image with its $N_{max}$ closest neighbors.
In our implementation, $N_{max}$ is set to 50.
We use NetVLAD~\cite{netvlad} to get retrieval results, and there is no limitation to using other image retrieval methods.

When there are too few initial feature matches, it is difficult to find enough candidate pairs for sufficient expansion.
To alleviate the deficiency of the initial connection, we add extra candidates according to the vote of the retrieval results.
If an image has many similar images in the registered image set, it is more likely to be registered.
Based on this, we perform feature matching on the images whose retrieval results contain many images in $A_{appr}$. 

On the other hand, the initial retrieval results of sequential images are often to be limited to adjacent images, which sometimes leads to ignoring the loop.
In order to solve this problem, we use the initial feature matches to cluster the images, and select a representative frame from each cluster to form a representative image set.
In the iterative matching process, only the representative image set participates in the calculation.
The other images are matched between the associated clusters after the iterative matching.
The overlap of adjacent representative images is small, so the retrieval results are not concentrated in adjacent images, which makes the loop easier to be found.
Moreover, using representative images for iterative matching also improves the speed of the algorithm.

The proposed iterative matching method improves speed but requires constructing a covisibility graph to describe image association, which incurs additional memory overhead. 
We store the covisibility graph in adjacency list form, with space complexity $O(|V| + |E|)$, which is linearly related to the number of vertices and edges.
Assuming the number of input image is $N_I$, the number of vertices in the covisibility graph is $N_I*N_p^2$, where $N_p$ is a constant. 
We denote the average number of covisible images per image as $N_c$.
The number of edges is proportional to $N_I*N_c$. 
In the worst case, $N_c = N_I$ and the space complexity is $O(N_I^2)$ but this rarely occurs which requires all image pairs to be covisible.
In general, $N_c$ is independent of the size of the scene and depends on the density of the camera distribution.
Based on our observations, $N_c$ ranges from 100 to 400 in most datasets, allowing us to achieve linear space complexity.
 
\section{Reconstruction Stage}
This section presents the reconstruction stage of the proposed SfM system. 
The traditional reconstruction stage has two main modules: the estimation module and the optimization module.
The estimation module performs the registration of frames and generates map points from feature tracks.
The optimization module performs local BA and low-frequency global BA to jointly optimize cameras and map points by minimizing the reprojection error.
We make the following improvements.
First, a novel error correction method detects the geometric error after each image registration and tries to correct the large error to enhance the robustness and keep the global consistency of the scene reconstruction result.
Second, the hierarchical structure is used to represent the registration dependency for both sequential and unordered images, and a keyframe selection scheme based on the hierarchical structure is proposed.
In the following content, we describe these two improvements in detail.

\subsection{Error Correction}
In traditional incremental SfM, the estimation error will inevitably accumulate due to the symbiotic relationship between points and camera poses. 
Many incremental reconstruction systems rely on frequent global optimization to alleviate error accumulation, but this approach has a defect.
Once most 2D-3D correspondences are considered as outliers during frame registration, the bundle adjustment method may not eliminate errors effectively in the absence of sufficient observation.
This situation is common in sequential image sequences with loops, where the accumulated error prevents the loop from being closed. 
SLAM systems solve this problem by explicitly closing loops~\cite{strasdat2010scale,mur2017visual}, but rely on known image order. 
In order to process arbitrary data, we propose a geometric error detection method and error correction module that explicitly closes loops without relying on image order.

\noindent\textbf{Error detection. }
Measuring the geometric error of registration is an important part in maintaining the accuracy of registration.
The traditional SfM systems~\cite{rome,lSfM,colmap} evaluate the quality of registration by the reprojection error, but the reprojection error relies on good 2D-3D correspondences.
When the 2D-3D correspondences are considered as outliers due to accumulated error, the reprojection error becomes too unreliable to evaluate the quality of frame registration.
ENFT-SfM~\cite{enft} employs the gradient direction of reprojection error to reduce cumulative error with a coarse-to-fine optimization, which also relies on good initial correspondences.
The relative motion estimation between frames by decomposing the essential matrix is another way to measure registration quality.
It doesn't rely on 3D points but is easily affected by feature matching noise and dynamic targets. 
And beyond that, when the camera is in pure rotation or the scene is a planar structure, the decomposition of the essential matrix will degrade.
Epipolar error is an effective indicator to evaluate the quality of two-view reconstruction, which measures the distance from a point to the epipolar line. 
However, the epipolar error is defined in the image domain, measured in pixels.
It is not straightforward to deduce the 6DoF registration error from the epipolar error, such as how many meters of translation error or how many degrees of rotation error.
Therefore, we re-formulate the epipolar geometry and deduce the registration error.

\begin{figure}
\centering
\includegraphics[width=0.6\linewidth]{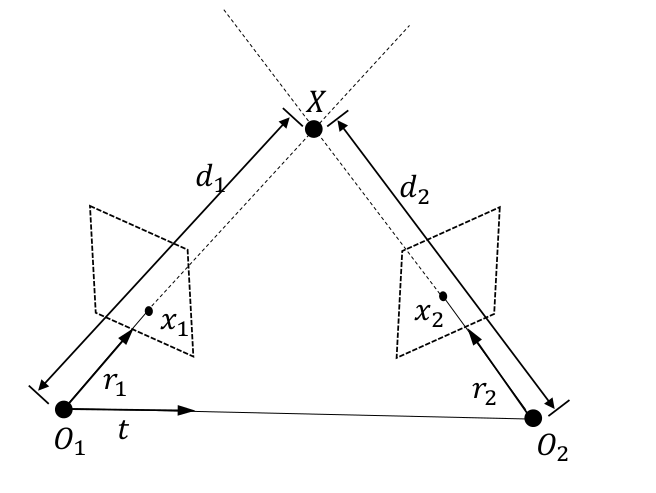}
\caption{\textbf{Two-view geometry}. The ray $r_1$ from the center of the camera center $O_1$ and the ray $r_2$ from the center of the camera center $O_2$ are compared to a 3D point $X$.
}
\label{fig:geometry1}
\end{figure}
\begin{figure}
\centering
\includegraphics[width=0.8\linewidth]{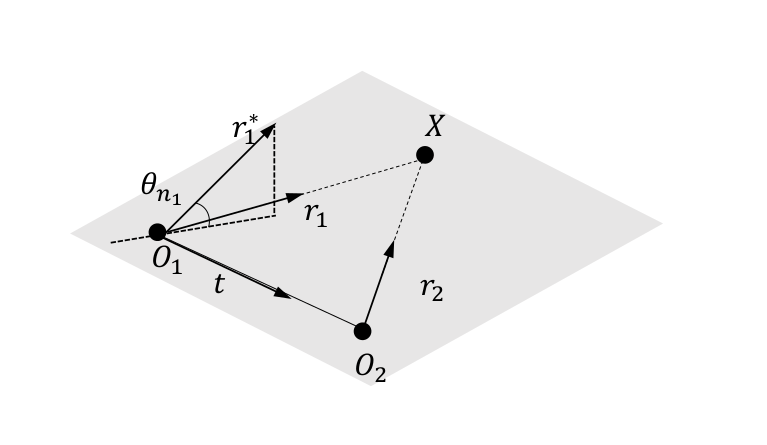}
\caption{The geometry of the errors in unit vector $r_1$. $r_1^*$ is the actual estimated value. $\theta_{n_1}$ is the angle of the $r_1^*$ and the plane $O_1 O_2 X$. }
\label{fig:geometry2}
\end{figure}

In order to better illustrate the derivation process, we now turn to a simple case that two images $I_1$ and $I_2$ observe the same 3D point $X$, and $x_1$,$x_2$ are the corresponding observations in normalized coordinates. 
The pose of $I_1$ is $(R_1,t_1)$ and the pose of $I_2$ is $(R_2,t_2)$.
Without considering the noise, the equation of $x_1^TEx_2 = 0$ should be satisfied, where $E$ is the essential matrix derived from the two poses.
In our method, we also start from the classic two-view geometry but considering the influence of each error term.
As Fig.~\ref{fig:geometry1} shows

\begin{equation}
t_1 + d_1 r_1 = t_2 + d_2 r_2 \label{eq:1}
\end{equation} 
where $d_i$ means the length between the camera center $O_i$ and $X$, $r_i$ means the unit vector in line $X - O_i$.
To remove depth factors, we cross product (\ref{eq:1}) by $r_2$ and dot by $r_1$.
\begin{equation}
((t_2 - t_1)\times{r_2})\cdot r_1 = (d_1 r_1\times{r_2}) r_1 - (d_2 r_2\times{r_2}) r_1  = 0 \label{eq:2}
\end{equation}

The above derivation is very common in the work related to the essential matrix. Actually, this equation is another representation of $x_1^TEx_2 = 0$, because $r_i = R_i\frac{x_i}{|x_i|} $.
In order to better explore the physical meaning, then we rewrite the formula (\ref{eq:2}) to

\begin{equation}
\frac{t\times r_2}{|t\times r_2|}\cdot r_1 = 0\label{eq:index}
\end{equation}
where $t = \frac{t_2-t_1}{|t_2-t_1|}$.
From a geometric point of view, $\frac{t\times r_2}{|t\times r_2|}$ represents the normal of the plane $O_1 O_2 X$ and the value of $\frac{t\times r_2}{|t\times r_2|}\cdot r_1$ is equal to $sin(\theta)$ that $\theta$ is the angle of the ray $r_1$ and the plane $O_1 O_2 X$. 
Without registered error and noise, this value is equal to 0. 

Next, we consider the effect of the various error on (\ref{eq:index}).
For simplicity, we consider the case where noise is added to the coordinates of the first image only. 
In order to better understand the influence of each error, first of all, we only consider the error of $r_1$.
we denote $r_1^*$ is the estimated value that $r_1^* = r_1 + n_1$,

\begin{equation}
\begin{aligned}
\frac{t\times r_2}{|t\times r_2|}\cdot r_1^* &= \frac{t\times r_2}{|t\times r_2|}\cdot n_1 \\
&= sin(\theta_{n_1})\cdot |n_1| \leq |n_1|
\end{aligned}
\end{equation}

As shown in Fig.~\ref{fig:geometry2}, $\theta_{n_t}$ is the angle of the $r_1^*$ and the plane $O_1 O_2 X$. Then we consider the error of $t$, and denote $t^*$ as the estimated value that $t^* = (t + n_t)$. 

\begin{equation}
\begin{aligned}
\frac{t^*\times r_2}{|t^*\times r_2|}\cdot r_1 &= \frac{t\times r_2+n_t\times r_2}{|t^*\times r_2|}\cdot r_1 \\ 
&= \frac{n_t\times r_2\cdot r_1 }{|t^*\times r_2|}\\
&= \frac{n_t\times r_2\cdot r_1 }{sin(\theta_{n_t})}\leq \frac{|n_t|}{sin(\theta_{n_t})} 
\end{aligned}
\end{equation}
where $\theta_{n_1}$ is the angle of the $t^*$ and the $r_2$.
Comprehensive consider the error of $r_1^*$ and $t^*$, we have
\begin{equation}
\begin{aligned}
\frac{t^*\times r_2}{|t^*\times r_2|}\cdot r_1^* &= \frac{t\times r_2+n_t\times r_2}{|t^*\times r_2|} \cdot (r_1+n_1) \\ &= \frac{n_t\times r_2\cdot r_1 }{|t^*\times r_2|} + \frac{t^*\times r_2\cdot n_1 }{|t^*\times r_2|} \\ &\leq \lambda |n_t|+|n_1|
\end{aligned}
\end{equation}
here $\lambda = \frac{1}{sin(\theta_{n_t})}$.
For convenience, we denote $V_{error}$ as the value of $\frac{t^*\times r_2}{|t^*\times r_2|}\cdot r_1^*$.
We prove that $V_{error}$ has a clear geometry significance and reflects the deviation of a relative pose.
By evaluating $V_{error}$, we can find the pairs of cameras with large relative position errors.
Similar to the essential matrix, $V_{error}$ does not encapsulate the scale. 
We introduce estimated depth to alleviate this problem.
For a given relative pose, the point depth can be estimated from feature matching.
If the estimated depth has a large error with the existing 3D points or exceeds a reasonable value, it reflects the potential poor relative pose.

In the actual reconstruction process, when the frame $I_i$ is registered, we detect the geometric relationship between $I_i$ and elements in $S_i$.
$S_i$ is a set of registered frames that has a matching relationship with $I_i$.
For each frame pair in ${(I_i,I_j)|I_j \in S_i}$, we compute $V_{error}$ of every feature matches.
Assuming that a good registration satisfies $|n_t|<a$ and $|n_1|<b$, then $V_{error} > \frac{a}{\lambda} + b$ will reflect a potential bad registration.
Considering there are some mismatches and small dynamic objects, only if the proportion of bad feature matches is large, we think there is a structure error.

\noindent\textbf{Error correction. }
When we detect large geometric errors, it usually means that 2D-3D correspondences are not good.
The most significant situation is that, due to the cumulative error, the newly registered frame is hardly associated with the other end of the loop, and the same region of the scene is reconstructed into two parts.
SLAM systems solve this problem by specially processing the loops to establish sufficient 2D-3D correspondences, but they rely on sequential images.
In previous SfM systems, merging 3D points and re-triangulation can alleviate this problem to a certain extent, but still can not deal with excessive cumulative errors.
We propose an error correction method that can deal without relying on the image order.

After each registration, we first detect geometric errors.
Specifically, for a newly registered frame $I_i$, we refer to the set of all registered frames that have a matching relationship with $I_i$ as $S_i$.
Then, for every image pair consisting of $I_i$ and one frame in $S_i$, we apply the above error detection and divide $S_i$ into two parts $S_i^1$ and $S_i^2$ that $S_i^1$ includes the part with the correct geometry and $S_i^2$ includes the rest.
In most cases, $S_i^2$ is an empty set, which means that the local map near $I_i$ is consistent.
In other cases, the local map has two separate parts, $S_i^1$ and $S_i^2$.
Similar to the loop closure strategy in the SLAM system, we register the current frame with the part local map of $S_i^1$ and $S_i^2$ respectively, and get two camera poses and the inlier 2D-3D correspondences.
Generally, bundle adjustment can eliminate these errors with all correspondence.
When starting from bad initial values, the optimization method will fall into the local optimum.
In this case, we use pose graph optimization to get a better initial value.

\subsection{Hierarchy-based Keyframe Selection}

\begin{figure}[t]
\centering
\includegraphics[width=0.9\linewidth]{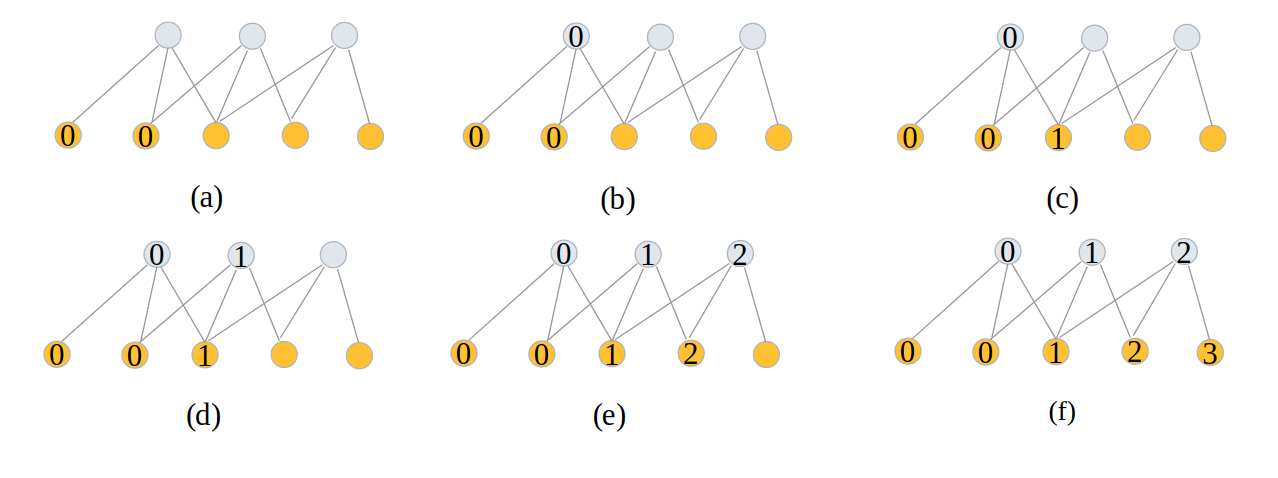}
\caption{\textbf{Steps in the calculation hierarchy process}.
The yellow nodes represent cameras and the gray nodes represent the landmarks.
Starting with two initial cameras in (a), the level of one landmark is set to zero, because it observed two cameras  whose levels are zero in (b).
Then the level of a camera is set to one in (c) ($N$ is 1 in this simple example).
Finally, all the levels of nodes are set.}
\label{fig:level_process}
\end{figure}

\begin{figure}[t]
\centering
\includegraphics[width=1.0\linewidth]{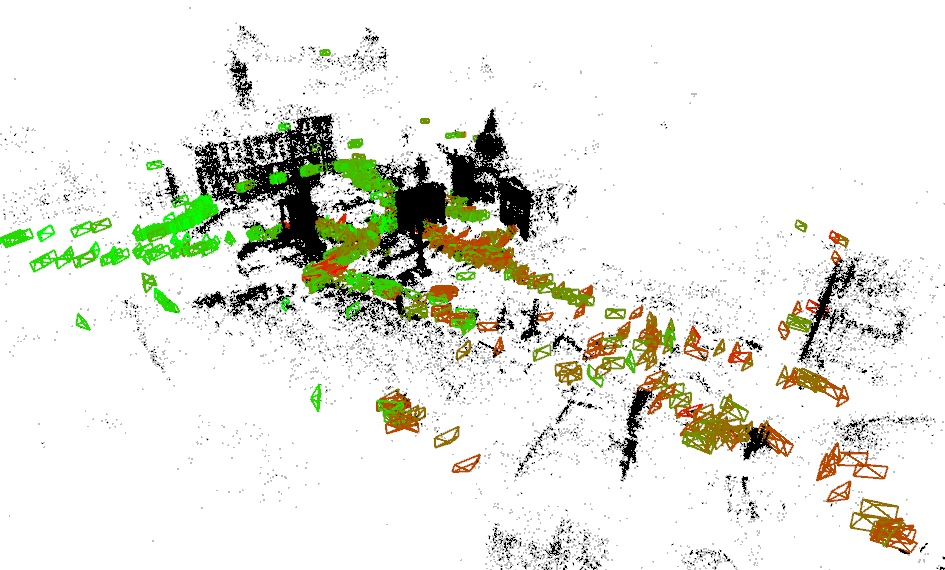}
\caption{\textbf{Visualization of the hierarchical structure}.
It is the reconstruction result of Roman Forum.
Cameras were colored from low level (green) to high level (red).
The reconstruction is initialized from the left, so the level of frames gradually increases from left to right. 
}
\label{fig:level}
\end{figure}

In order to improve the reconstruction speed, we propose a hierarchy-based keyframe selection method.
Keyframe strategy is widely employed in SLAM systems, and skeleton graphs~\cite{snavely2008skeletal} and icon images~\cite{li2008modeling, frahm2010building} are similar methods in the field of SfM.
The core idea of these methods is to reduce the amount of computation by reducing the number of images actually involved in the reconstruction.
Specifically, a keyframe set that can express the whole scene is extracted from all images for reconstruction.
In this way, the complexity of the reconstruction is reduced to the complexity of the scene itself rather than the number of images.
 
The selection of keyframes is relatively easy in sequential images.
It only needs to ensure sufficient matches between two adjacent keyframes, so that deleting the intermediate image between the two keyframes will not affect the registration of other frames.
For unordered data, the image association is complex, so the selection of keyframes becomes difficult.
Therefore, we judge deleting which images will not affect the registration of other frames instead of directly judging which images are keyframes.
The images that do not affect the registration of other images after being deleted are called redundant frames.
For finding redundant frames, it is very important to recover the registration dependency between frames.
We propose a novel hierarchical structure, which can effectively represent the frame registration dependencies in an arbitrary scene.
The hierarchy of previous reconstruction methods\cite{hierarchical,gao2022irav3} are mainly used to divide images into several clusters, and then reconstruct and merge them layer by layer to obtain complete reconstruction results efficiently. 
Unlike them, the hierarchy we propose is only used to describe the image correlation in any data, and only affects the selection of key frames, without affecting other reconstruction modules.

\noindent\textbf{Hierarchical structure.}
We propose a simple hierarchical generation method.
The hierarchical levels of the initial two frames are set to 0. 
The points which are observed by at least two frames of level ($0 \sim n$) are set to level $n$.
This ensures that the points of level $n$ can be triangulated with the frames of level ($0 \sim n$).
Similarly, the frames which observe at least 50 points of level ($0 \sim n$) are set to level $n + 1$.
This ensures that the frames of level $n+1$ can be registered with the points of level ($0 \sim n$).
In this way, all frames and points are assigned to different levels.
In order to better illustrate the generation of the hierarchy, we show a simple example in Fig.~\ref{fig:level_process}.
On the one hand, this hierarchical structure reflects the registration dependency.
The estimation of SfM variables (camera poses and map points) at a high level depends on the variables at low levels.
On the other hand, the hierarchical relationship also implies the "distance" from the initial two frames. 
The estimation of a high-level variable often has higher uncertainty, because they are farther from the initial frames.
Fig.~\ref{fig:level} shows the visualization of the hierarchical structure on a real dataset.

\noindent\textbf{Keyframe selection.}
Through the registration dependency described by a hierarchical structure, we design a convenient and fast redundant frame detection method.
For each frame $I$, we compute the level $n_I$ of frame $I$ and record the number $m_I$ of points of level ($0 \sim n_I$) that frame $I$ can observe.
Assuming that $I_i$ is deleted, we first calculate the new hierarchical level of points that $I_i$ can observe.
Then, we calculate a new $m_{I_j}$ for each matched frame $I_j$ of $I_i$.
Once the new $m_{I_j}$ is smaller than 50, it indicates that deleting $I_i$ will change the hierarchical level of $I_j$, and in this case, $I_i$ is considered to be a keyframe.
Otherwise, $I$ is considered as a redundant frame.
Even if we delete all redundant frames, each keyframe can observe at least 50 3D points, which ensures that they can still be registered.
Moreover, these remaining keyframes maintain the same hierarchical level as the original, which means that they are not farther away from the initial frames. 

Back to the implementation of the keyframe algorithm, a list of keyframes was maintained in the reconstruction stage.
In the beginning, all newly registered frames will be added to this keyframe list.
And before each global optimization, we will check the keyframe list and remove redundant frames.
In the optimization process, we only adjust the poses of keyframes and the 3D points they can observe, which greatly reduces the amount of calculation in global bundle adjustment.
Besides, the poses of the redundant frames are modified by the change of 3D points after optimization.

\section{Experimental Results}

\begin{center}
	\begin{table*}[tbh!]
	\caption{Evaluation Results on 14 Large-scale Unordered Internet Photo Collections.}
	\resizebox{1\linewidth}{!}{
	\begin{tabular}{lrrrrrrrrrrrrrrrrrrrl}
		\hline
		& \textbf{\#Size} && \multicolumn{4}{c}{\textbf{\#Registered}} && \multicolumn{4}{c}{\textbf{\#Time[s]}} && \multicolumn{4}{c}{\textbf{\#Precision[\%]}}&& \multicolumn{3}{c}{\textbf{\#Recall[\%]}} \\
		& && $IR_{5}$ & $IR_{25}$ & $IR_{50}$ & $Ours$ && $IR_{5}$ & $IR_{25}$ & $IR_{50}$ & $Ours$ && $IR_{5}$ & $IR_{25}$ & $IR_{50}$ & $Ours$&& $IR_{5}$ & $IR_{25}$  & $Ours$ \\
		\hline
		Alamo & 2,915 & & 683 & 810 & 862 & 760 & & 144 & 722 & 1405 & 233 & & 40.74 & 27.61 & 23.68 & 37.29            && 87.70 & 95.79 & 95.95 \\
		Ellis Island & 2,587 & & 295 & 344 & 351 & 331 & & 104 & 584 & 988 & 177 & & 49.25 & 33.27 & 26.94 & 48.47      && 59.33&93.03&95.11\\
		Gendarmenmarkt & 1,463 & & 702 & 984 & 1020 & 923 & & 62 & 346 & 596 & 230 & & 59.16 & 46.25 & 39.86 & 52.09    && 57.04&91.69&94.38\\
		Madrid Metropolis & 1,344 & & 245 & 409 & 435 & 406 & & 47 & 244 & 430 & 113 & & 42.31 & 27.76 & 23.10 & 37.35   && 72.49&91.86&93.74\\
		Montreal Notre Dame & 2,298 & & 475 & 554 & 564 & 552 & & 99 & 523 & 972 & 171 & & 54.79 & 41.97 & 36.31 & 48.54&& 76.09&96.82&97.83\\
		NYC Library & 2,550 & & 385 & 614 & 574 & 592 & & 102 & 544 & 975 & 170 & & 44.32 & 28.60 & 21.74 & 43.58       && 64.47&92.95&93.83\\
		Piazza del Popolo & 2,251 & & 332 & 901 & 951 & 865 & & 99 & 468 & 872 & 234 & & 47.10 & 34.15 & 28.53 & 43.23  && 65.46&90.16&91.25\\
		Piccadilly & 7,351 & & 2213 & 2871 & 2988 & 2838 & & 406 & 1508 & 2717 & 995 & & 40.95 & 29.17 & 24.73 & 40.63  && 54.92&81.43&86.36\\
		Roman Forum & 2,364 & & 1291 & 1500 & 1599 & 1546 & & 164 & 587 & 1226 & 473 & & 59.45 & 43.15 & 35.57 & 33.87  && 74.90&94.02&94.07\\
		Tower of London & 1,576 & & 477 & 651 & 699 & 632 & & 84 & 386 & 732 & 199 & & 42.65 & 28.08 & 22.41 & 35.58    && 70.93&94.95&94.64\\
		Trafalgar & 15,685 & & 4397 & 7048 & 7725 & 7122 & & 713 & 3474 & 6396 & 2819 & & 41.41 & 30.76 & 26.54 & 37.98 && 52.28&81.00&88.19\\
		Union Square & 5,961 & & 536 & 985 & 1070 & 971 & & 311 & 1436 & 2313 & 449 & & 22.63 & 13.48 & 10.29 & 30.07   && 78.96&96.08&78.96\\
		Vienna Cathedral & 6,288 & & 924 & 1060 & 1119 & 1033 & & 533 & 1657 & 3328 & 707 & & 40.25 & 25.67 & 20.69 & 39.28 && 99.31&99.59&99.31 \\
		Yorkminster & 3,368 & & 452 & 655 & 1060 & 927 & & 165 & 1182 & 1620 & 382 & & 48.79 & 32.09 & 24.89 & 37.16    && 72.55&93.59&93.81\\	
		\hline
		Average & 4,142 & & 957 & 1384 & 1501 & 1392 & & 216 & 975 & 1755 & 525 & & 45.27 & 31.57 & 26.09 & 40.37       &&70.46&92.35&92.67 \\	
		\hline 
	\end{tabular}
	}
	\label{table:matching}
\end{table*}
\end{center}

We evaluate our algorithm on several real datasets, including Internet photo collection~(1DSfM dataset~\cite{1dSfM}), vehicle-loaded videos (KITTI dataset~\cite{geiger2013vision}), a set of handheld videos~\cite{enft}, and a set of Internet city walking tour videos.
The Internet city walking tour videos are downloaded from YouTube.
The experiments are conducted on a desktop PC with an Intel i7-9700K 3.6GHz CPU, 64GB of memory, and a NVIDIA GTX 2070 graphics card.
The experiments are divided into two parts.
First, we quantitatively evaluate the efficiency of the matching strategy and the influence of relevant parameters on the matching result.
Then, we compare our reconstruction part with the state-of-the-art and carry out the ablation experiments to verify the effectiveness of the proposed error correction and hierarchy-based keyframe selection algorithm.

\subsection{Matching Stage}

\begin{figure}[t]
\centering
\includegraphics[width=0.9\linewidth]{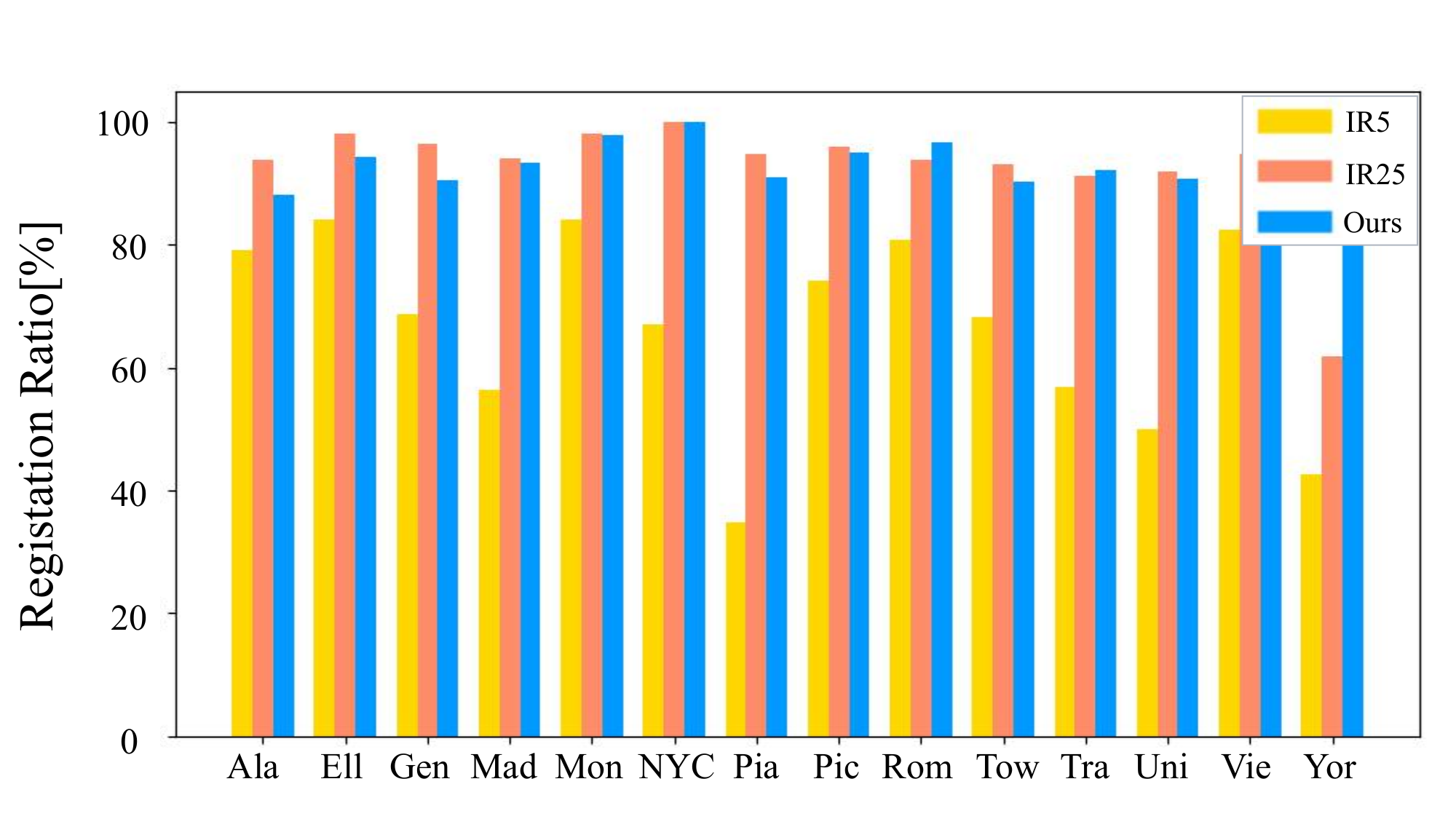}
\caption{The registration ratio of different methods. The registration ratio is generated from dividing the number of registered frames by $Registered_{IR_{50}}$.}
\label{fig:reg_ratio}
\end{figure}

\begin{figure}[t]
\centering
\includegraphics[width=0.9\linewidth]{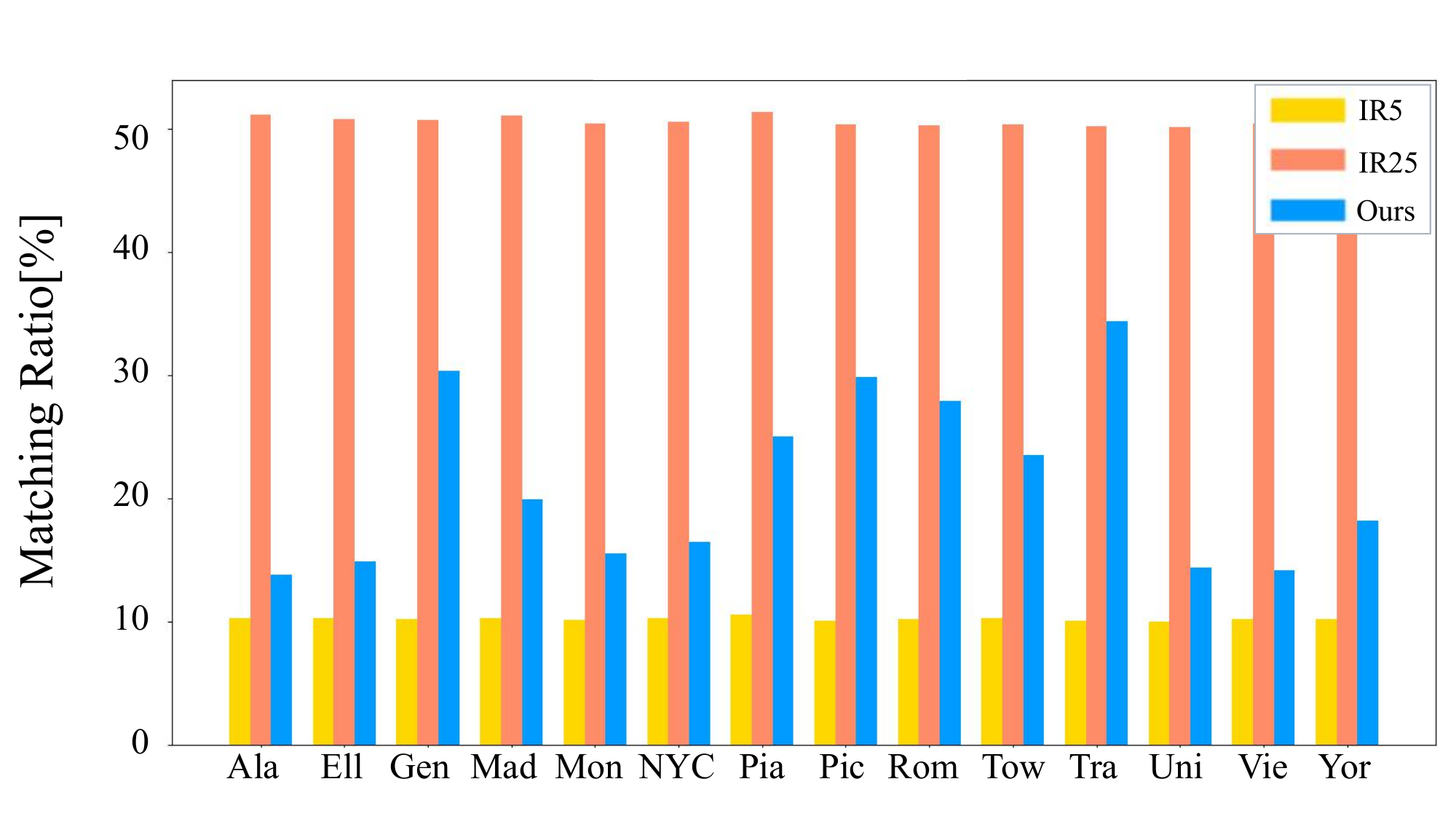}
\caption{The number of matching operations for different methods.
To plot the results of different methods on the same coordinate system, we normalize the results by dividing the number of matching operations for each method with the number of matching operations for $IR_{50}$.}
\label{fig:mat_ratio}
\end{figure}

In this section, we conduct comparison experiments on unordered and sequential datasets to verify the superiority of the proposed method in terms of efficiency, and evaluate the impact of each component as well as the influence of different parameters.

\begin{figure*}[t]
\centering
\includegraphics[width=0.65\linewidth]{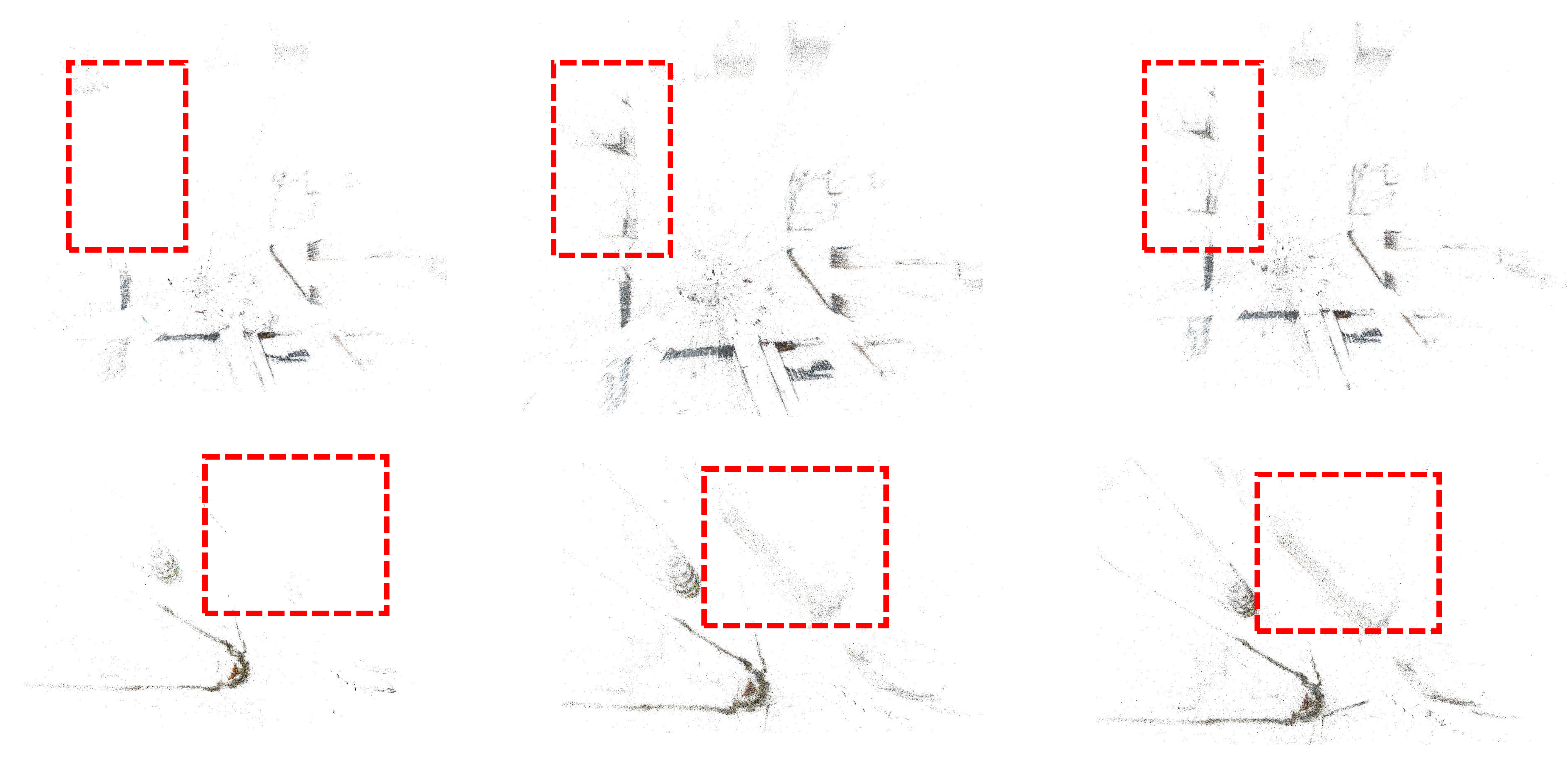}
\caption{The top view of reconstruction results.
The first row is the reconstruction result of Madrid Metropolis,
the second row is the reconstruction result of Union Square.
The left column is $IR_{5}$, the middle column is $IR_{50}$, the right column is $Ours$. 
We highlight the area which shows the differences in reconstruction completeness of different methods.
}
\label{fig:reconstruction}
\vspace{-1.0em}
\end{figure*}

For unordered images, we conduct the experiments on 14 large unordered datasets~\cite{1dSfM}.
These 14 datasets contain a total number of 58k unordered Internet photos, covering a wide variety of scenes.
The strategy which retrieves $N_R$ images to match is denoted as $IR_{N_R}$.
We use the state-of-the-art image retrieval method NetVLAD~\cite{netvlad}, and compare $IR_{5}$, $IR_{25}$, $IR_{50}$, and the proposed matching strategy.
In the proposed strategy, the two inputs of Algorithm~\ref{alg:iteration}, $ \mathcal{M}_{init}$ is the matching result of $IR_{5}$ and $N_{max}$ is set to 50.
When $N_{max}$ is 50, the result of the proposed strategy happens to be a subset of the result of $IR_{50}$. 
For a fair comparison, all strategies use the same implementation in feature extraction, feature matching, and geometric verification.
We evaluate four metrics and the results are shown in Table~\ref{table:matching}.
$Registered$ is the number of registered frames,
$Time$ denotes the time consumed in the matching stage, 
$Precision$ is computed from $\frac{TP}{P}$,
and $Recall = \frac{TP}{N_{gt}}$, 
where $TP$ is the number of image pairs that share at least 30 feature tracks, 
$P$ is the number of the candidate image pairs,
$N_{gt}$ is the true number of covisible images. 
It is best to obtain $N_{gt}$ by brute-force matching, but it is computationally prohibitive for large-scale datasets.
Considering that the matching results of all strategies are subsets of the result of $IR_{50}$, we can use the result of $IR_{50}$ as $N_{gt}$ without affecting the comparison result between different strategies.
Note that when calculating recall, matches that are irrelevant to the registered frames are removed.
$IR_{5}$ is the fastest method, but we find the results of $IR_{5}$ can not guarantee the completeness of reconstruction. 
Fig.~\ref{fig:reconstruction} shows the reconstruction results of the different methods in Madrid Metropolis and Union Square.
In order to better show the completeness of reconstruction results under different matching methods, we show the registration ratio ($\frac{Registered}{Registered_{IR_{50}}}$) in Fig.~\ref{fig:reg_ratio}.
Compared with $IR_5$, the reconstruction result of our method is more complete.
Compared with $IR_{25}$ and $IR_{50}$, our method has a comparable registration ratio but is significantly faster.
The improvement in speed comes from the reduction in matching operations. 
We plot the number of matching operations for different methods to demonstrate the advantages of the proposed method in Fig.~\ref{fig:mat_ratio}.
The proposed method reduces the number of matching operations by $30\sim 60\%$ compared to $IR_{25}$, while maintaining a similar number of registered frames.
On the one hand, the reduction in the number of match operation comes from the improvement of precision.
The precision of our method is higher than $IR_{25}$ and $IR_{50}$ on average, which means that our method can accurately predict overlapping images.
On the other hand, our method only matches the potentially registered frames, while the image retrieval based matching method matches all frames.
This difference also causes a gap in speed.
In addition, compared with $IR_{5}$ and $IR_{25} $, the recall of our method is the highest on average.

\begin{table}[t!]
\centering
\caption{Evaluation Results on KITTI odometry datasets.} 
\begin{tabular}{lrrrrrrrrrl}
	\hline 
	&\multicolumn{2}{c}{\textbf{\#Time[s]}} 
	&\multicolumn{2}{c}{\textbf{\#Precision[\%]}}
	&\multicolumn{2}{c}{\textbf{\#Recall[\%]}} \\
	&  $Seq.$ & $Ours$ &  $Seq.$  & $Ours$ &  $Seq.$  & $Ours$   \\
    \hline 
	00 &  677 & 429 & 54.67 & 55.65 &100.00&99.98 \\
	01 &  100 & 75  & 59.45 & 64.23 &100.00&99.98 \\
	02 &  907 & 453 & 51.20 & 55.56 &100.00&99.98  \\
	03 &  144 & 127 & 61.14 & 57.38 &100.00&100.00 \\
	04 &  36 & 28   & 61.57 & 57.88 &100.00&99.98  \\
	05 &  430 & 352 & 57.81 & 57.56 &99.99&99.85    \\
	06 &  163 & 151 & 61.95 & 53.42 &99.85&99.89    \\
	07 &  149 & 164 &  58.65 & 63.87 &100.00&99.98  \\
	08 &  690 & 535 &  53.53 & 46.39 &100.00&99.99  \\
	09 &  246 & 162 &  51.39 & 45.17 &100.00&100.00 \\
	10 &  181 & 174 &  55.11 & 50.93 &100.00&100.00 \\
	\hline 
\end{tabular} 
\label{table:seq_match}
\end{table}

To prove that the proposed method can handle the sequential images as well, we compare our matching method with the sequential matching method on KITTI dataset.
Since both our method and this sequential matching method can register all images, we do not compare the number of registered frames, but mainly compare the other three metrics, and the results are shown in Table~\ref{table:seq_match}.
The sequential matching method is the implementation of COLMAP and is denoted as $Seq$. 
In the sequential matching of COLMAP, each frame is matched with 10 adjacent frames, and the image-retrieval-based matching is performed every 10 frames.
It can be found that our method can process sequential images, and the speed of our method is slightly faster for most sequences, with comparable precision. 
It is worth noting that our method does not need to know whether the input data set is sequential in advance, which is a great advantage over the traditional matching algorithm.

\begin{table}[t]
	\centering
	\caption{The Results of Different $N_p$ on Yorkminster.}
	\begin{tabular}{|l|r|r|r|r|r|}
	\hline
		 & \textbf{Reg.} &  \textbf{Num.} &  \textbf{Time[s]} & \textbf{Pre.[\%]}& \textbf{Rec.[\%]} \\ \hline
        $N_p=1,T=2$  & 960 & 34714 & 482  & 27.11& 94.24 \\ \hline
	$N_p=5,T=2$  & 953 & 31651 & 439  & 31.84& 94.09 \\ \hline
	$N_p=10,T=2$ & 943 & 29220 & 405  & 34.46& 93.99 \\ \hline
	$N_p=20,T=2$ & 927 & 27752 & 382  & 37.16& 93.81 \\ \hline
	$N_p=30,T=2$ & 618 & 23046 & 320  & 39.32& 91.11  \\ \hline
	$N_p=40,T=2$ & 601 & 21653 & 302  & 42.19& 91.09  \\ \hline
	\end{tabular}
	\label{N_p} 
\end{table}

\begin{table}[t]
		\centering
		\caption{The Results of Different $T$ on Yorkminster.}
		\begin{tabular}{|l|r|r|r|r|r|}
		\hline
			 & \textbf{Reg.} & 
            \textbf{Num.} &  
             \textbf{Time[s]} & 
             \textbf{Pre.[\%]}& 
             \textbf{Rec.[\%]} \\ \hline
			 $N_p=5,T=1$ & 962  &34732  & 486 &  29.52&94.67    \\ \hline
			 $N_p=5,T=2$ & 953  &31651  & 439 &  31.84&94.09    \\ \hline
			 $N_p=5,T=4$ & 921  &26646  & 372 &  36.96&93.05    \\ \hline
			 $N_p=5,T=6$ & 573  &17134  & 228 &  40.61&89.79    \\ \hline
\end{tabular}
\label{T} 
\end{table}

\begin{table}[t]
	 \centering
	\caption{The Experiments Results for $A_{appr}$ on Ellis Island.}
	\begin{tabular}{|l|r|r|r|r|r|}
	\hline
	& \textbf{Reg.} & \textbf{Num.}& \textbf{Time[s]} & \textbf{Pre.[\%]}& \textbf{Rec.[\%]} \\ 
	\hline
	with $A_{appr}$    & 331    & 14730& 177& 59.80&95.11\\ \hline
	without $A_{appr}$ & 335    & 22483& 264& 52.14&95.99\\ \hline
	\end{tabular}
	\label{A_t}
\end{table}

$N_p$ controls the number of patches in each image and $T$ denotes the minimum common tracks between a covisible patch pair. 
To show the effects of the parameters $N_p$ and $T$, we present the ablation studies on Yorkminster in Table~\ref{N_p} and Table~\ref{T}.
We show the number of matching operations for different methods to display the impact of different parameters more intuitively.
Increasing $N_p$ and $T$ reduce the time consumption and raise the precision but decrease the number of registered images.  
Since increasing $N_p$ means an image is divided into more small cells, it skips many mismatched image pairs, so the precision is high.
Similarly, a strict requirement for $T$~($T=6$) also has high precision.
However, at the same time, many potential matches are ignored and not found, resulting in fewer registered images.
We found the proposed method can keep a good balance between speed and registered number with $N_p \in [5,20]$ and $T \in [2,4]$.
In our implementation, $N_p$ is set to 20, and $T$ is set to 2 for efficiency.

To evaluate the improvement of predicting potential registration frames $A_{appr}$, we present the experiments with and without computing $A_{appr}$ on the dataset Ellis Island.
As listed in Table~\ref{A_t}, the registered image number of the method with $A_{appr}$ is almost the same as that of the method without $A_{appr}$, but the running time is reduced by 1/3.

\subsection{Reconstruction Stage}

\begin{center}
	\begin{table*}[tbh!]
	\caption{Evaluation Reconstruction Results for State-of-the-art SfM Systems on Large-scale Photo/Video Collections.} 
	\resizebox{1\linewidth}{!}{
	\begin{tabular}{lrrrrrrrrrrrrrl}
		\hline
		& \textbf{\#Size} && \multicolumn{3}{c}{\textbf{\#Registered}} && \multicolumn{3}{c}{\textbf{\#Time[s]}} && \multicolumn{3}{c}{\textbf{\#Avg. Reproj. Error [px]}} \\
		& && $Theia$  & $COLMAP$ & $Ours$ &
		& $Theia$ & $COLMAP$ & $Ours$ &
		& $Theia$ & $COLMAP$ & $Ours$ \\
		\hline
		Garden & 9971 & 
		        & 2855 &  9955 & 9955 &
		        & 3124 &  38367 & 5410 &
		        & 1.64 &  0.57 & 0.60 \\%
        Roman Forum MIX & 5227 & 
		        & 2140 &  3158 & 4005 &
		        & 582 &  18159 & 1358 &
		        & 1.26 &  0.71 & 0.54 \\%
		\hline
		Alamo & 2,915 & 
		        & 799 &  882 & 815 &
		        & 326 &  1259 & 136 &
		        & 0.73 &  0.66 & 0.66 \\
		Ellis Island & 2,587 & 
        		& 322 &  344 & 343 &
        		& 173 &  258 & 31 & 
        		& 0.91  & 0.78 & 0.76 \\
		Gendarmenmarkt & 1,463 & 
        		& 962 &  974 & 930 &
        		& 173 &  1877 & 162 & 
        		& 0.99  & 0.68 & 0.69 \\
		Madrid Metropolis & 1,344 & 
		        & 391  & 412 & 406 & 
		        & 90  & 810 & 40 & 
		        & 0.78  & 0.61 & 0.63 \\
		Montreal Notre Dame & 2,298 & 
		        & 549  & 555 & 552 & 
		        & 286  & 973 & 103 & 
		        & 0.99  & 0.82 & 0.79 \\
		NYC Library & 2,550 & 
		        & 586  & 611 & 556 & 
		        & 316  & 786 & 62 & 
		        & 0.89  & 0.69 & 0.69 \\
		Piazza del Popolo & 2,251 & 
		        & 885  & 909 & 895 & 
		        & 130 & 1276 & 93 & 
		        & 0.84  & 0.67 & 0.68 \\
		Piccadilly & 7,351 & 
        		& 2691 & 2871 & 2862 & 
        		& 263  & 4403 & 634 & 
        		& 1.18 & 0.74 & 0.80 \\
		Roman Forum & 2,364 & 
		        & 1451  & 1499  & 1496 &
		        & 311   & 2726  & 208  &
		        & 1.05  & 0.71 & 0.66 \\
		Tower of London & 1,576 & 
		        & 625  & 648 & 644 &
		        & 196  & 1156 & 67 & 
		        & 0.73 & 0.62 & 0.64 \\
		Trafalgar & 15,685 & 
		        & 6610 & 7083 & 7022 & 
		        & 534 & 11467 & 2819 & 
		        & 1.03 & 0.71 & 0.72 \\
		Union Square & 5,961 & 
		        & 852  & 996 & 962 & 
		        & 51  & 920 & 40 &
		        & 1.22  & 0.67 & 0.70 \\
		Vienna Cathedral & 6,288 & 
		        & 1043  & 1055 & 1046 & 
		        & 474  & 1957 & 189 & 
		        & 0.80  & 0.72 & 0.72 \\
		Yorkminster & 3,368 & 
		        & 633  & 661 & 649 & 
		        & 335 & 1278 & 130 &
		        & 0.89  & 0.71 & 0.69 \\	
		\hline 
	\end{tabular}
	}
	\label{table:unorder}
\end{table*}
\end{center}
 
\begin{table}[tbh]
	\caption{Translation RMSE and Time Comparison on KITTI Odometry Dataset.}
	\resizebox{1\linewidth}{!}
	{\begin{tabular}{lrrrrr}
		\hline
		&\multicolumn{3}{c}{\textbf{\#RMSE}}&\multicolumn{2}{c}{\textbf{\#Time[s]}}\\
		& \tiny{$ORB-SLAM3$} & \tiny{$COLMAP$} & \tiny{$Ours$}   & \tiny{$COLMAP$} & \tiny{$Ours$}\\
		\hline
		00  &7.28&52.04&\textbf{4.95}    &11988&\textbf{600}\\ 
		01  &X&8.12&\textbf{7.08}        &1761&\textbf{59}\\ 
		02  &\textbf{21.50}&53.32&23.35 &12473&\textbf{515}\\ 
		03  &1.59&1.68&\textbf{1.33}     &1530&\textbf{77}\\ 
		04  &1.40&0.68&\textbf{0.63}     &329&\textbf{17}\\ 
		05  &5.29&16.44&\textbf{4.23} &7438&\textbf{245}\\ 
		06  &13.50&36.90&\textbf{3.86}   &2688&\textbf{94}\\ 
		07  &\textbf{2.26}&22.74&3.60   &1818&\textbf{185}\\ 
		08  &46.68&124.77&\textbf{44.00} &7899&\textbf{433}\\ 
		09  &\textbf{6.62}&60.86&7.79   &2600&\textbf{132}\\ 
		10  &8.80&12.25&\textbf{5.37}   &2323&\textbf{135}\\ 
		\hline
	\end{tabular}
	}
	\label{table:kitti}
\end{table}

\begin{table}[tbh]
	\caption{Translation RMSE and Time Comparison for Different Strategies on KITTI Odometry Dataset.} 
	\resizebox{1\linewidth}{!}
	{\begin{tabular}{lrrrrrr}
		\hline
		&\multicolumn{3}{c}{\textbf{\#RMSE}}&\multicolumn{3}{c}{\textbf{\#Time[s]}}\\
		  & $gba$ & $kgba$  & $kgba+$  & $gba$ & $kgba$ & $kgba+$         \\
		\hline
		00&40.51&28.74  &\textbf{4.95}    &2130 & 349 & 600   \\
		01&12.08&7.08   &\textbf{7.08}     &459  & 59  & 59        \\
		02&28.27&27.52  &\textbf{23.35}  &3588 & 439 & 515     \\
		03&1.48 &1.33   &\textbf{1.33}      &243  & 75  & 77   \\
		04&1.20 &0.63   &\textbf{0.63}      &39   & 17  & 17   \\
		05&25.26&18.02  &\textbf{4.23}    &1296 & 213 & 245  \\
		06&28.71&32.92  &\textbf{3.86}    &355  & 68  & 94   \\
		07&15.51&18.59  &\textbf{3.60}    &589  & 164  & 185   \\
		08&84.79&93.81  &\textbf{44.00}   &1571 & 389  & 433   \\
		09&77.33&77.61  &\textbf{7.79}    &570  & 108  & 132   \\
		10&6.81 &5.37   &\textbf{5.37}      &571  & 132  & 135   \\
		\hline
	\end{tabular}
	}
	\label{table:kgba}
\end{table}

\begin{table}[tbh]
	\caption{Time Comparison for Different Strategies on 1DSfM Dataset.} 
	\resizebox{1\linewidth}{!}
	{\begin{tabular}{lrrrrrr}
		\hline
		&\multicolumn{3}{c}{\textbf{\#Registered}}&\multicolumn{3}{c}{\textbf{\#Time[s]}}\\
& $gba$ & $kgba$  & $kgba+$  & $gba$ & $kgba$ & $kgba+$  \\
		\hline
    Alamo               &872    &815    &815    &739  & 134  & 136  \\
    Ellis Island        &351    &343    &343    &181  & 31   & 31       \\
    Gendarmenmarkt      &974   &930    &930     &871  & 160  & 162     \\
    Madrid Metropolis   &408    &406    &406    &497  & 40   & 40   \\
    Montreal Notre Dame &553    &552    &552    &838  & 101  & 103   \\
    NYC Library         &571    &556    &556    &171  & 61   & 62  \\
    Piazza del Popolo   &910    &895    &895    &291  & 92   & 93   \\
    Piccadilly          &2889   &2862   &2862   &3297 & 628  & 634   \\
    Roman Forum         &1501   &1496   &1496   &1001 & 206  & 208  \\
    Tower of London     &645    &644    &644    &384  & 67   & 67  \\
    Trafalgar           &7042   &7022   &7022   &13942 & 2810 & 2819   \\
    Union Square        &1027   &962    &962    &270  & 40   & 40  \\
    Vienna Cathedral    &1141   &1046   &1046   &1268 & 185  & 189   \\
    Yorkminster         &656   &649    &649     &580  & 125  & 130   \\
		\hline
	\end{tabular}
	}
	\label{table:kgba_uno}
\end{table}

In order to verify the efficiency of the reconstruction process, we conducted experiments on the unordered dataset~\cite{1dSfM}, KITTI dataset~\cite{geiger2013vision}, and two complex datasets.
The Rome MIX dataset consists of Internet images and YouTube videos, and the Garden data set consists of six video sequences.
These datasets contain various scenes and different lighting and angles.
The optimizer of our system is the open-source solver Ceres~\cite{ceres}.

\begin{figure}
\centering
\includegraphics[width=1.0\linewidth]{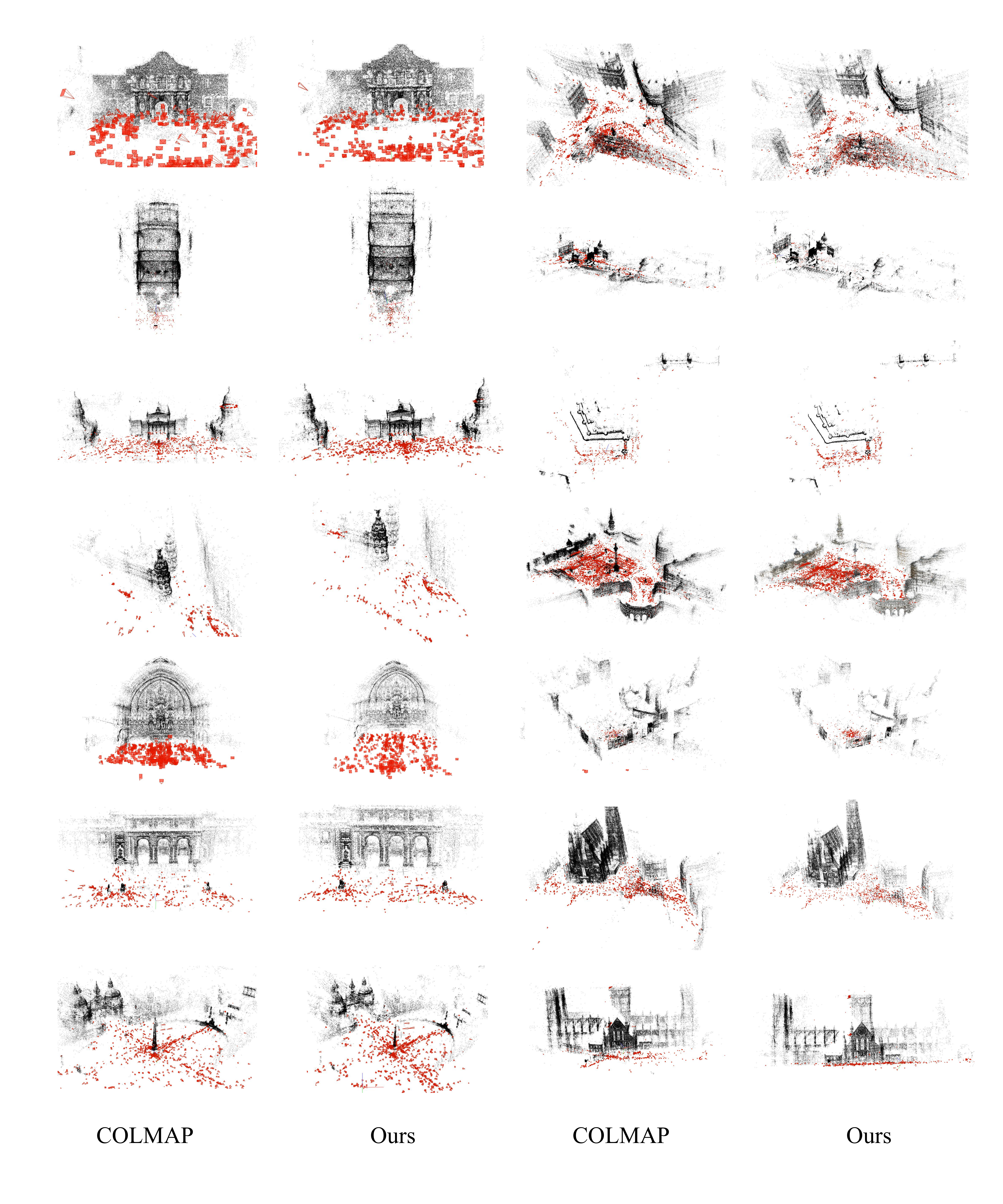}
\caption{Reconstruction results of different methods on unordered datasets.}
\label{fig:compare_1dsfm}
\end{figure}

For the unordered dataset and mixed dataset, we compare the proposed reconstruction part to the state-of-the-art incremental SfM system COLMAP~\cite{colmap} and the global SfM system Theia~\cite{theia-manual}.
Throughout all experiments, we use the same feature matching result as input, and compare the reconstruction results of different SfM systems. 
The comparison results are shown in Table~\ref{table:unorder}.
For each data, we compare the result of the largest reconstruction component.
Theia, as a global SfM, is the fastest method on large datasets (Trafalgar, Piccadilly, Garden, and Roman Forum MIX).
However, due to weakness in handling outlier, it generates poor or incomplete reconstruction structures on these large datasets.
The reprojection error of our system is lower than Theia, indicating the superior accuracy of our method.
Thanks to keyframe-based GBA, our reconstruction speed is even faster than Theia on most datasets that are relatively small.
Both our method and COLMAP are incremental methods, but our method is an order of magnitude faster than COLMAP with a comparable number of registered frames and reprojection error.
The reconstruction results of our method and COLMAP on 1DSfM dataset are shown in Fig.~\ref{fig:compare_1dsfm}.

\begin{figure}[t]
\centering
\includegraphics[width=1.0\linewidth]{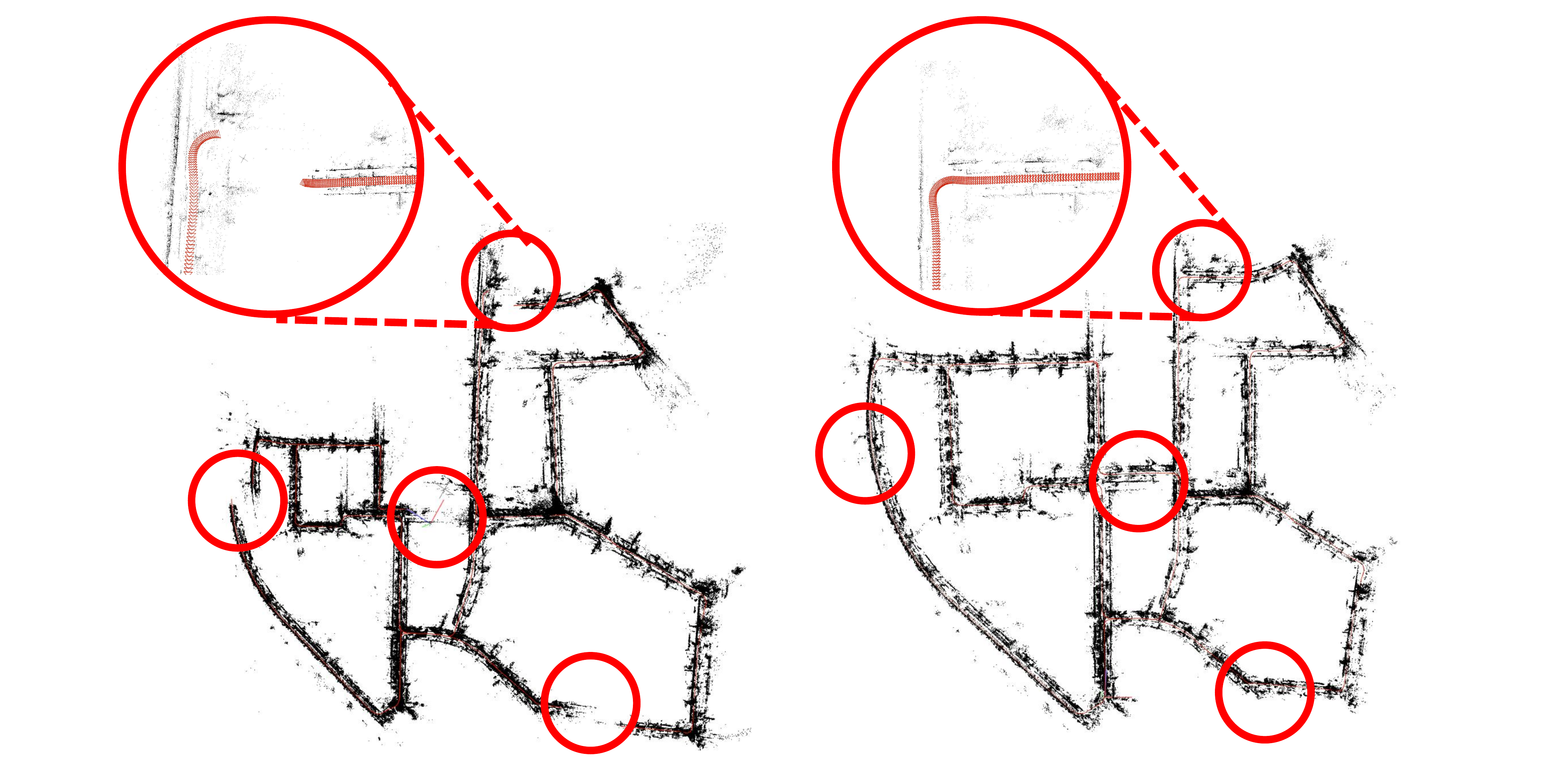}
\caption{The comparison between the reconstruction results of COLMAP (left) and ours (right) on sequence 00. We use red circles to highlight the differences. }
\label{fig:comparison_kitti}
\end{figure}

To compare the accuracy of different methods, we evaluate ORB-SLAM3\cite{orbslam3}, COLMAP, and our system in the KITTI dataset with the groundtruth poses obtained by high precision GPS/IMU.
To ensure the fairness of the comparison, we fixed the camera intrinsics parameters, and only use the monocular image sequences.
The results are shown in Table~\ref{table:kitti}.
ORB-SLAM3 runs at ten frames per second and carried out matching and mapping at the same time as a multi-threaded system.
It is difficult to obtain the mapping time alone, so Table~\ref{table:kitti} does not show the reconstruction time of ORB-SLAM3.
Compared with ORB-SLAM3, our method has better accuracy because of sufficient optimization and more feature correspondences.
Although there is still a gap with the SLAM system in speed (our system has not reached ten frames per second), as a general system, our system can handle unordered and mixed data well with good efficiency.
Furthermore, some sequences (00, 02, 05, 06, 07, 08, 09) contain loops.
In these sequences, COLMAP system has large errors without closing the loop,
while the proposed error correction deals with loop scenes well.
As shown in Fig.~\ref{fig:comparison_kitti}, there are several places of misalignment by COLMAP highlighted by red circles, while our scene structure has a good global consistency.
It is worth noting that the error correction module does not need to know the order of input data, which is different from the traditional sequential loop closure algorithm.
Moreover, compared to unordered datasets, the sequential datasets are more redundant, so the efficiency gain of the proposed method compared to COLMAP is more impressive.

To demonstrate the effectiveness of the proposed keyframe-based global bundle adjustment and the geometric error correction, we compare the reconstruction result with three strategies on the KITTI dataset and 1DSfM dataset.
The strategy $gba$ performs global optimization in all frames,
$kgba$ use the proposed keyframe-based global optimization,
and $kgba+$ reconstructed scene with both the keyframe-based global bundle adjustment and the geometric error correction. 
For sequential data, we evaluate the accuracy of camera trajectory and reconstruction time.
Table~\ref{table:kgba} shows $kgba$ greatly accelerates the reconstruction speed compared to $gba$.
And in most sequences, the trajectory errors of $kgba $ and $gba$ are similar.
Due to the lack of loop closure, $gba$ and $kgba$ have scale drift in some sequences, which makes the error large.
$kgba+$ solves this problem well by closing loops, improving the reconstruction accuracy at a good speed.
For unordered data, there is no ground-truth of camera poses, so we evaluate the number of registered images to show the completeness of reconstruction results.
Table~\ref{table:kgba_uno} shows the acceleration of $kgba$ is equally effective on unordered data.
$kgba+$ is mainly used to handle loop scene and has little effect on unordered data or sequential data without loops (seq.01, seq.03, seq.04, seq.10).

\section{Conclusion}
This paper proposes an efficient SfM, which can deal with both unordered and sequential data in a unified framework.
The proposed SfM system can effectively predict the covisibility by some existing feature matches to extend feature matching, so as to accelerate the matching stage.
A hierarchy-based keyframe-selection method is proposed to improve the reconstruction speed.
The comprehensive evaluation shows that the matching speed of the proposed SfM is three times that of the state-of-the-art, and the reconstruction speed has an order of magnitude advantage over the excellent incremental SfM system COLMAP.
Our future work will explore how to integrate the learning-based method with the proposed method to achieve a more efficient and robust SfM system.

\section*{Acknowledgment}
The authors would like to thank Zhaopeng Cui, Bangbang Yang, and Yijin Li for their help in proofreading the paper.

\ifCLASSOPTIONcaptionsoff
\newpage
\fi

\bibliographystyle{IEEEtran}
\bibliography{jrnl}

\begin{IEEEbiography}
[{\includegraphics[width=1in,height=1.25in,clip,keepaspectratio]{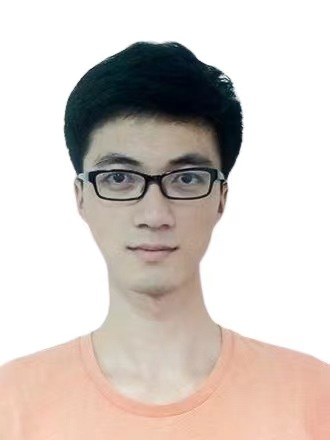}}]
{Zhichao Ye}
received the B.S. degree from Shandong University in 2017 and his Ph.D. degree in computer science from Zhejiang University in 2022.
He is currently a Researcher with SenseTime Research.
His current research interests include structure-from-motion, SLAM, augmented reality.
\end{IEEEbiography} 
\vspace{-2.0em}
\begin{IEEEbiography}
[{\includegraphics[width=1in,height=1.25in,clip,keepaspectratio]{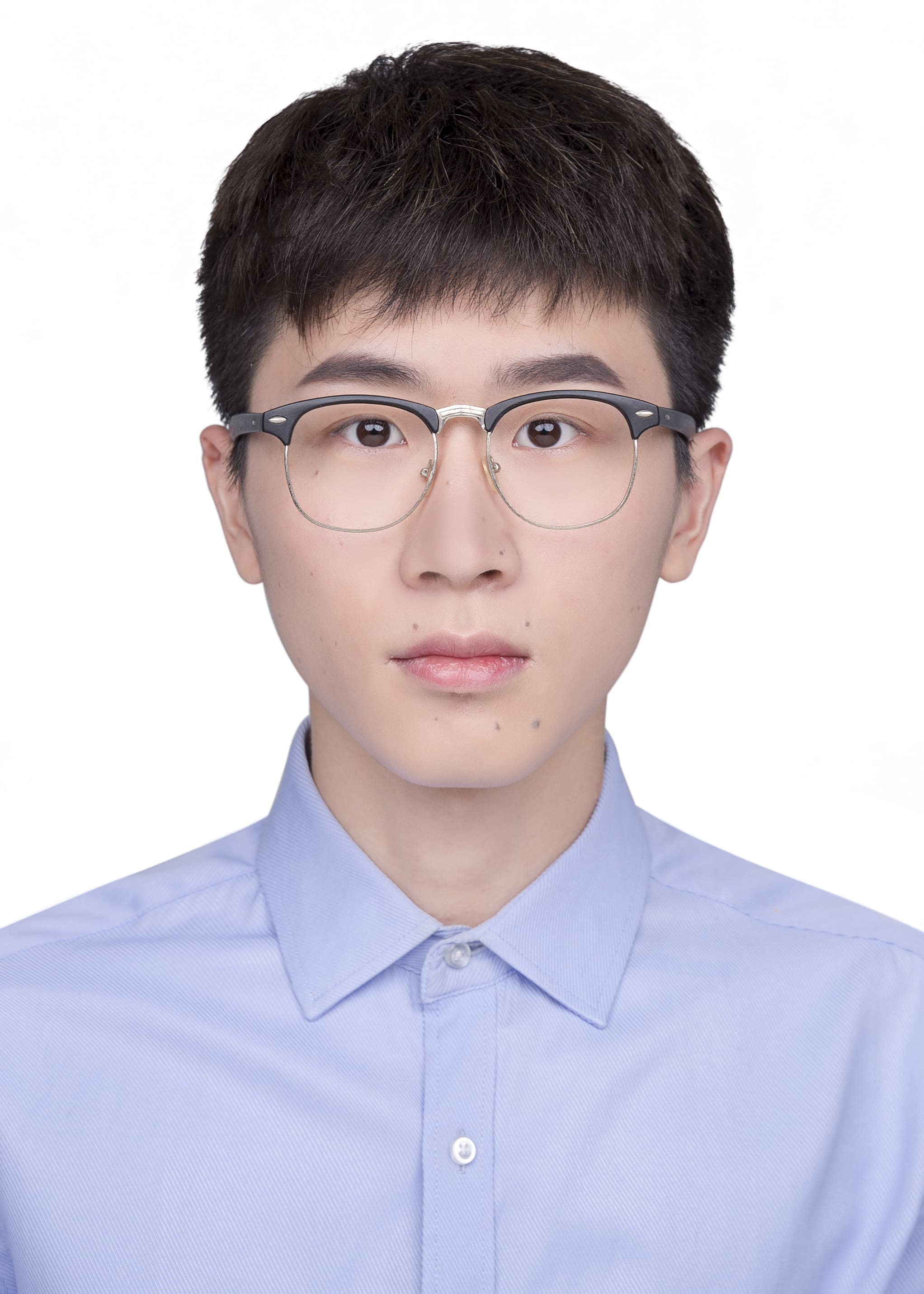}}]
{Chong Bao} received the B.S. degree from Zhejiang Gongshang University, China, in 2020. 
He is currently pursuing the Ph.D. degree with Zhejiang University.
His current research interests include 3D reconstruction and neural rendering.
\end{IEEEbiography} 
\vspace{-2.0em}
\begin{IEEEbiography}
[{\includegraphics[width=1in,height=1.25in,clip,keepaspectratio]{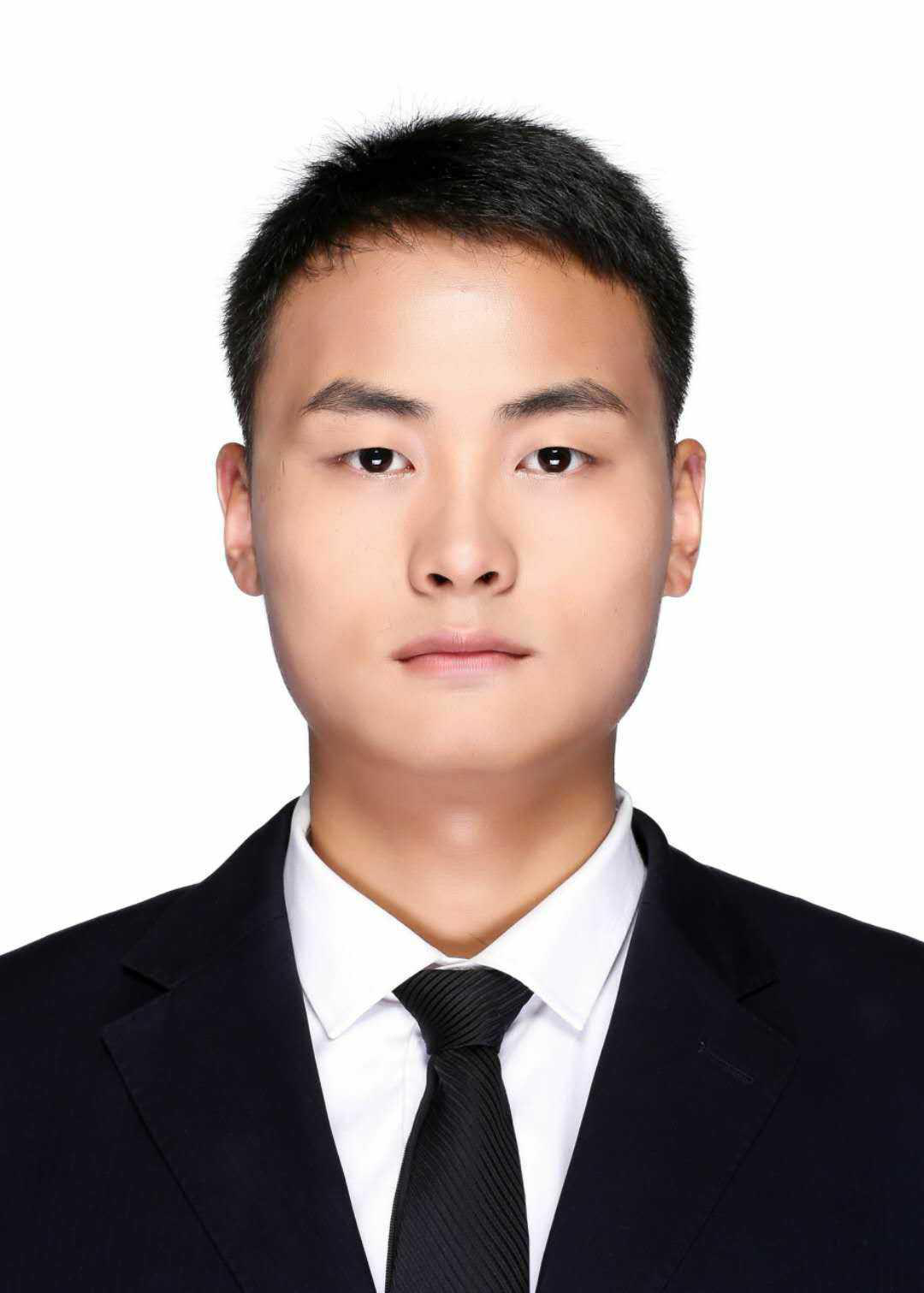}}]
{Xin Zhou}
 received his bachelor's degree in Computer Science from the University of Electronic Science and Technology of China in 2019 and the master's degree in computer science from Zhejiang University in 2022. His research interests include SLAM, 3D reconstruction, and autonomous driving.
\end{IEEEbiography} 
\vspace{-2.0em}
\begin{IEEEbiography}
[{\includegraphics[width=1in,height=1.25in,clip,keepaspectratio]{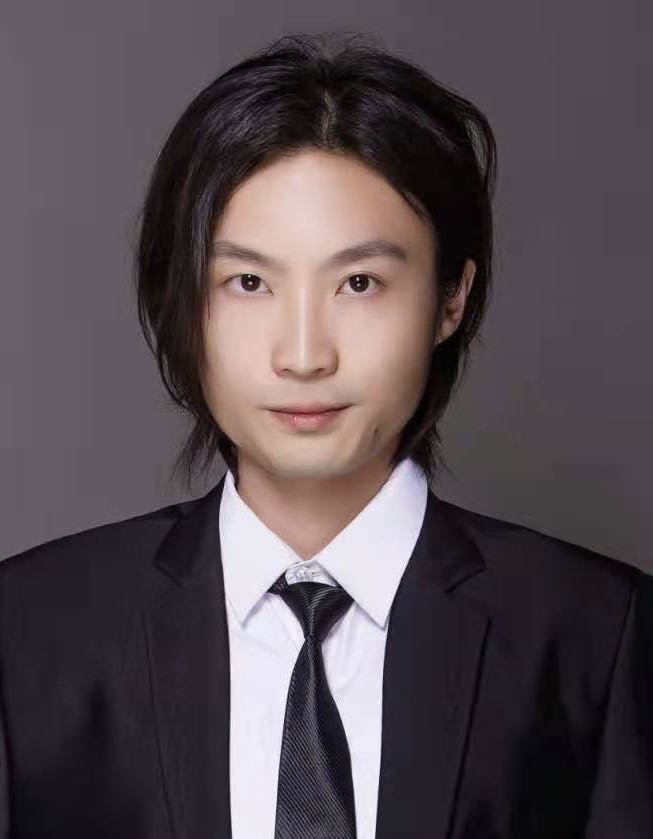}}]
{Haomin Liu} 
received the master's and Ph.D. degrees in computer science from Zhejiang University in 2009 and 2017. He is currently a Research Director of SenseTime Research. His research interests include structure-from-motion, SLAM, and augmented reality.
\end{IEEEbiography} 
\vspace{-2.0em}
\begin{IEEEbiography}
[{\includegraphics[width=1in,height=1.25in,clip,keepaspectratio]{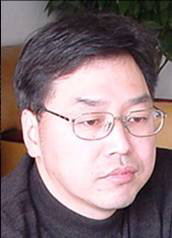}}]
{Hujun Bao} (Member, IEEE) is currently a professor in the Computer Science Department of Zhejiang University, and the former director of the State Key Laboratory of Computer Aided Design and Computer Graphics. His research interests include computer graphics, computer vision and mixed reality. He leads the mixed reality group in the lab to make a wide range of research on 3D reconstruction and modeling, real-time rendering and virtual reality, realtime 3D fusion and augmented reality. Some of these algorithms have been successfully integrated into the mixed reality system SenseMARS.
\end{IEEEbiography} 
\vspace{-2.0em}
\begin{IEEEbiography}
[{\includegraphics[width=1in,height=1.25in,clip,keepaspectratio]{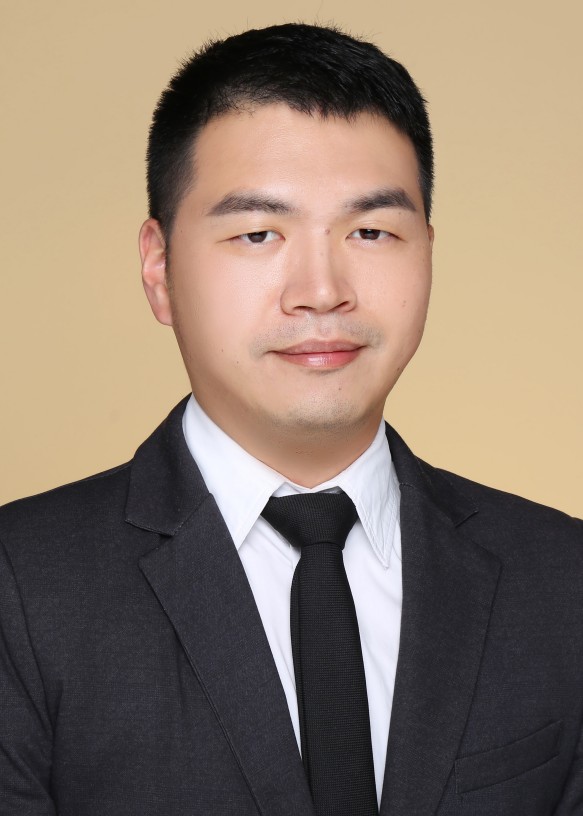}}]
{Guofeng Zhang} (Member, IEEE) is currently a professor at Zhejiang University. He received the B.S. and Ph.D. degrees in computer science and technology from Zhejiang University in 2003 and 2009, respectively. He received the National Excellent Doctoral Dissertation Award, the Excellent Doctoral Dissertation Award of China Computer Federation and the ISMAR 2020 Best Paper Award. His research interests include SLAM, 3D reconstruction and augmented reality.
\end{IEEEbiography}





\end{document}